\documentclass{article} 
\usepackage{arxiv,times}

\makeatletter
\def\@fnsymbol#1{\ensuremath{\ifcase#1\or \dagger\or \ddagger\or
   \mathsection\or \mathparagraph\or \|\or **\or \dagger\dagger
   \or \ddagger\ddagger \else\@ctrerr\fi}}
\makeatother

\usepackage{amsmath,amsfonts,bm}

\def\eqref#1{equation~\ref{#1}}

\def\1{\bm{1}}

\DeclareMathAlphabet{\mathsfit}{\encodingdefault}{\sfdefault}{m}{sl}
\SetMathAlphabet{\mathsfit}{bold}{\encodingdefault}{\sfdefault}{bx}{n}

\usepackage{graphicx}
\usepackage{wrapfig}
\usepackage{amsmath}
\usepackage{amssymb}
\usepackage{amsthm}
\usepackage{booktabs}
\usepackage{subcaption}
\usepackage{mathtools}
\usepackage{pifont}
\usepackage{cancel}
\usepackage{algorithm}
\usepackage{algorithmic}
\usepackage{multirow}
\usepackage{colortbl}
\usepackage{bbding}
\usepackage{utfsym}
\usepackage{xspace}
\usepackage{tabularx}
\usepackage{wrapfig}

\newtheorem{theorem}{Theorem}

\def\bff{\mathbf{f}}

\def\bfs{\mathbf{s}}

\def\bfu{\mathbf{u}}
\def\bfv{\mathbf{v}}
\def\bfw{\mathbf{w}}
\def\bfx{\mathbf{x}}

\def\bfG{\mathbf{G}}

\def\rmd{\mathrm{d}}

\def\bbE{\mathbb{E}}

\def\eqref#1{Eq.~(\ref{#1})}

\usepackage{hyperref}
\usepackage{url}

\title{Spectral Amplitude Purification in Distribution Matching for Diffusion Distillation}

\author{Zhenyu Zhou$^{1}$,\space\space\space
  Can Wang$^{1}$,\space\space\space
  Chun Chen$^{1}$,\space\space\space
  Zeyu Zheng$^{2}$,\space\space\space
  Defang Chen$^{2}$\thanks{Corresponding author.}
  \vspace{0.15cm}
  \\
  $^{1}$\normalfont{Zhejiang University}\quad
  $^{2}$\normalfont{University of California, Berkeley}
  \vspace{0.15cm}
  \\
  \tt\small
  {zhyzhou@zju.edu.cn,\quad
  defchern@berkeley.edu}
}

\iclrfinalcopy 
\begin{document}
\def\ourName{SAP-DMD\xspace}

\maketitle
\begin{abstract} 
Distribution Matching Distillation (DMD) enables high-quality diffusion sampling in only a few steps, but its optimization dynamics remain dominated by coarse, low-frequency signals, delaying the recovery of fine-grained details. 
We identify a pronounced concentration of spectral amplitudes at low frequencies in the DMD directional error, where dominant low-frequency components overwhelm weaker mid- and high-frequency signals. 
To address this issue, we propose \textbf{S}pectral \textbf{A}mplitude \textbf{P}urification for Distribution Matching Distillation (\ourName), a plug-and-play approach that adaptively modulates the amplitude spectrum of the DMD directional field. 
By suppressing the dominant tail of the amplitude spectrum, \ourName reduces low-frequency dominance and promotes more effective recovery of fine structures and textures. Experiments on PixArt-$\alpha$, SD3, and SD3.5 demonstrate that \ourName accelerates training convergence and improves generation quality under both 2-step and 4-step sampling.
\end{abstract}
\section{Introduction}
Diffusion models achieve remarkable generative quality across image~\citep{dhariwal2021diffusion,karras2022edm,rombach2022ldm,podell2024sdxl}, video~\citep{ho2022video,blattmann2023align,zheng2024open,seedance2026seedance}, and audio~\citep{kong2021diffwave,chen2021wavegrad,evans2025stable} generation, but typically require dozens of network evaluations during sampling~\citep{sohl2015deep,ho2020ddpm,song2021sde}. 
As modern generative models continue to scale~\citep{esser2024scaling,flux,wu2025qwen}, this iterative process incurs substantial computational overhead and latency. 
Few-step distillation addresses this bottleneck by compressing the sampling trajectory into only a few network evaluations~\citep{meng2023distillation,song2023consistency,yin2023one}. 
Among these approaches, Distribution Matching Distillation (DMD)~\citep{luo2023diff,yin2023one,yin2024improved} has emerged as a powerful paradigm that directly aligns the distributions of a few-step generator and a pretrained diffusion model.

\begin{figure}[t]
    \centering
    \begin{subfigure}[b]{0.475\textwidth}
        \centering
        \includegraphics[width=\textwidth]{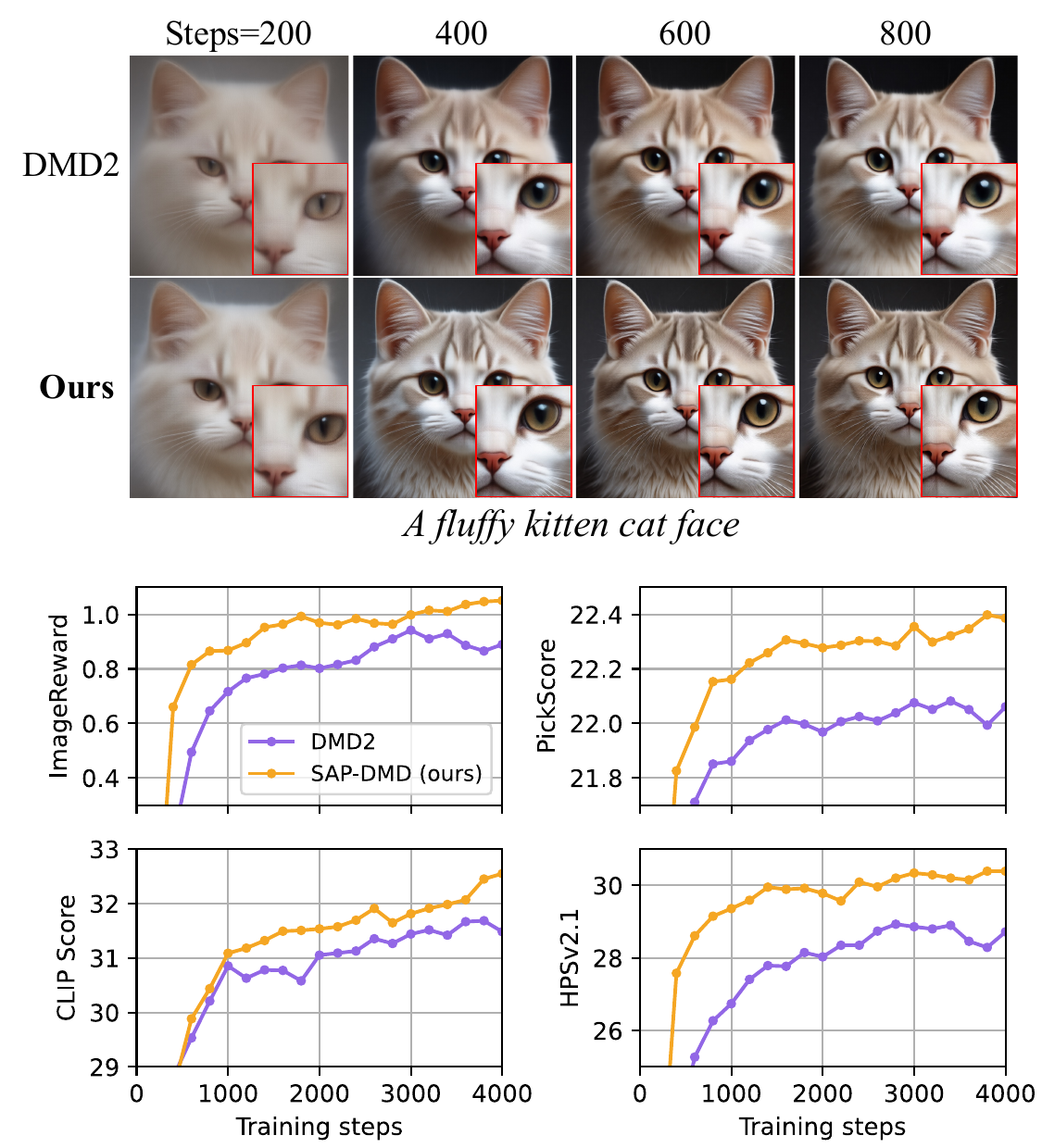}
        \caption{2-step performance evolution during training.}
        \label{fig:evolve}
    \end{subfigure}
    \hfill
    \begin{subfigure}[b]{0.515\textwidth}
        \centering
        \includegraphics[width=\textwidth]{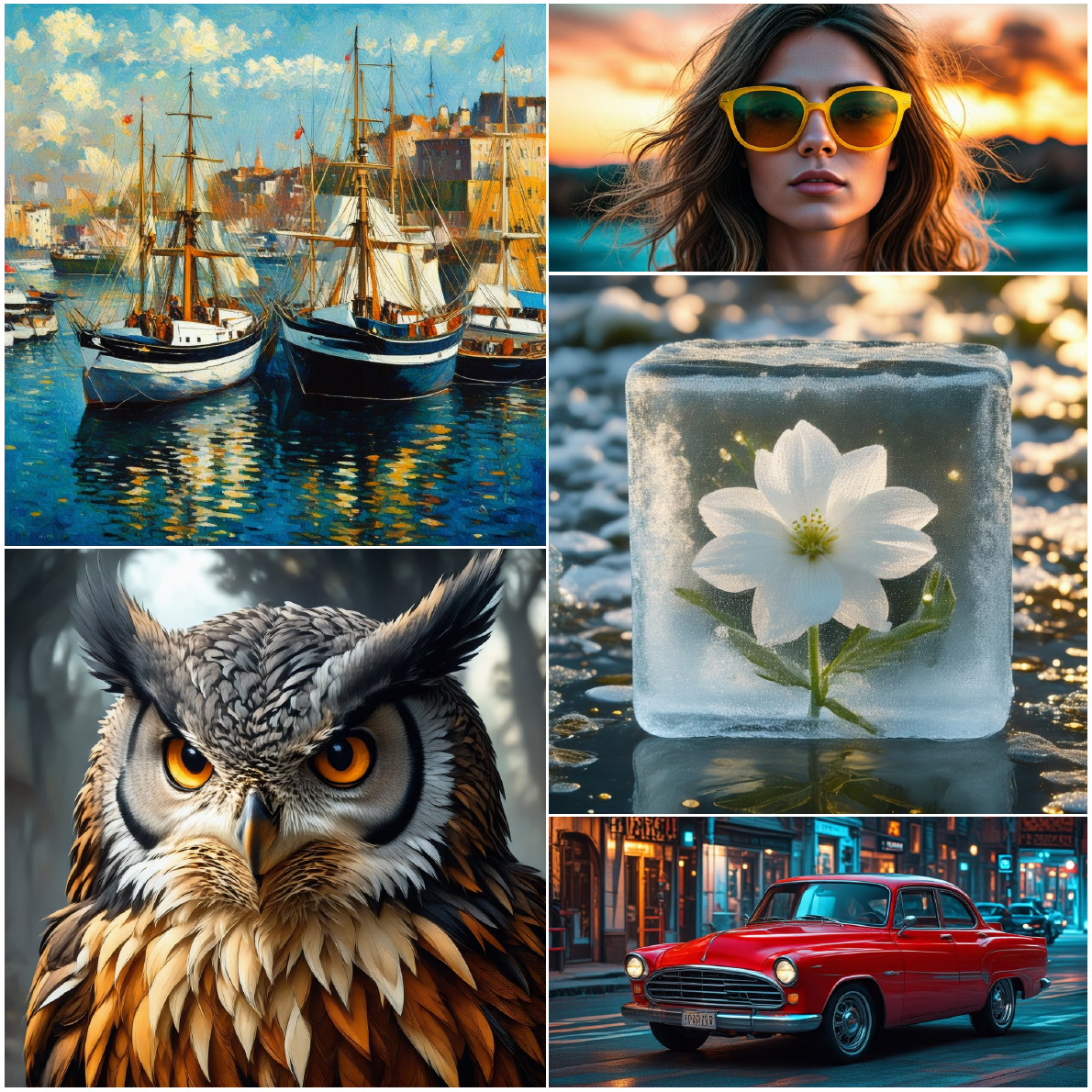}
        \caption{2-step qualitative samples.}
        \label{fig:teaser_qualitative}
    \end{subfigure}
    \caption{Performance overview on SD3.5 Medium.
    (a) Training evolution of DMD2 and \ourName. \ourName recovers fine details earlier and improves DrawBench~\citep{saharia2022photorealistic} evaluation metrics faster on all 200 prompts using 2-step generation without GAN loss. 
    (b) 2-step samples generated by \ourName, which show sharp structures and rich details across diverse prompts.
    }
    \label{fig:teaser}
\end{figure}

Despite the strong global fidelity of DMD, we observe that it recovers fine-grained details more slowly than coarse structures during training (see Figure~\ref{fig:teaser}). 
To understand this behavior, we analyze its optimization dynamics from a frequency-domain perspective. 
In DMD, the generator update is driven by the directional error between the real and fake model predictions. 
We find that its amplitude spectrum is strongly concentrated at low frequencies: a small number of low-frequency components dominate the directional signal, while substantially weaker mid- and high-frequency components contribute less to the update. 
Such low-frequency dominance can bias optimization toward coarse structures and delay the recovery of textures and sharp boundaries.

Motivated by this observation, we propose \textbf{S}pectral \textbf{A}mplitude \textbf{P}urification for Distribution Matching Distillation (\ourName), a plug-and-play approach that directly modulates the amplitude spectrum of the DMD directional field. \ourName adaptively suppresses large upper-tail spectral amplitudes while preserving the phase, thereby reducing the disproportionate influence of dominant spectral components without altering the underlying DMD framework. This simple modification promotes faster recovery of fine structures, requiring only lightweight FFT-based computation during training and no additional inference-time cost. Extensive experiments on PixArt-$\alpha$, SD3, and SD3.5 demonstrate clear improvements in training convergence and generation quality under both 2-step and 4-step sampling.

Our main contributions are summarized as follows:
\begin{itemize} 
    \item We reveal a spectral amplitude concentration in the DMD directional field, which is associated with under-emphasized fine-scale corrections and delayed fine-detail recovery. 
    \item We develop \ourName, a simple plug-and-play spectral purification method that adaptively suppresses excessive amplitudes while preserving phase and the distribution-matching fixed point.
    \item We demonstrate through extensive experiments on multiple text-to-image models that \ourName accelerates detail recovery and consistently improves few-step generation quality over competitive baselines. 
\end{itemize}
\section{Preliminaries}
\subsection{Diffusion Models}
Diffusion models connect the data distribution $q_{\text{real}}$ to a predefined noise distribution $q_{\text{noise}}$ through a stochastic forward process. Let $\bfx_t \in \mathbb{R}^{C \times H \times W}$ denote a sample at time $t \in [0,1]$. The forward process can be described by the stochastic differential equation (SDE) 
$\rmd \bfx_t = f(t)\bfx_t \rmd t + g(t)\rmd \bfw_t$, 
where $f(t)$ and $g(t)$ are the drift and diffusion coefficients, respectively, and $\bfw_t$ denotes standard Brownian motion~\citep{song2021sde}. Its transition kernel takes the form $q(\bfx_t | \bfx_0) = \mathcal{N}(\bfx_t; \alpha_t \bfx_0, \sigma_t^2 \mathbf{I})$, $\bfx_0 \sim q_{\text{real}}$, where $\alpha_t,\sigma_t \geq 0$ are time-dependent coefficients. For example, $\alpha_t^2+\sigma_t^2=1$ corresponds to variance-preserving diffusion~\citep{song2021sde}, while $\alpha_t=1-t$ and $\sigma_t=t$ define the linear interpolation used in flow models~\citep{lipman2022flow,liu2022flow}.

The forward SDE admits a corresponding probability flow ODE with the same marginal distributions $\{q_t\}_{t=0}^1$:
$\rmd \bfx_t = [f(t)\bfx_t - \frac{1}{2}g(t)^2 \bfs_\psi(\bfx_t, t)] \rmd t$,
where $\bfs_\psi(\bfx_t, t) \approx \nabla_{\bfx_t} \log q_t(\bfx_t)$ is the learned score function~\citep{hyvarinen2005estimation,vincent2011connection}. 
Flow-based models instead directly parameterize a deterministic velocity field:
$\rmd \bfx_t = \bfv_\psi(\bfx_t, t) \rmd t$.
Different prediction parameterizations can be related through the score function. In particular,
\begin{equation}
    \label{eq:parameterizations}
    \bfs_\psi(\bfx_t, t) 
    = -\frac{\bfx_t - \alpha_t \bff_\psi(\bfx_t, t)}{\sigma_t^2}, 
\end{equation}
where $\bff_\psi$ denotes the $\bfx_0$-prediction model. 
Under the linear flow parameterization $\alpha_t=1-t$ and $\sigma_t=t$, the velocity prediction $\bfv_\psi$ further satisfies $\bfs_\psi(\bfx_t, t) = -\frac{\bfx_t + \alpha_t \bfv_\psi(\bfx_t, t)}{\sigma_t}$. 
For simplicity, we omit the explicit dependence on $t$ when it is clear from context.

\subsection{Distribution Matching Distillation}
DMD trains a few-step generator $\bfG_\theta$ by minimizing the time-weighted reverse KL divergence between the generator marginal $p_{\theta,t}$ and the target marginal $q_t$~\citep{yin2023one}:
\begin{equation}
    \label{eq:loss_gen}
    \mathcal{L}_{\text{DMD}}(\theta) = \int_t \omega(t) \bbE_{\bfx_0\sim p_{\theta,0},\bfx_t \sim q(\bfx_t|\bfx_0)}\left[ \log p_{\theta,t}(\bfx_t) - \log q_t (\bfx_t) \right] \rmd t,
\end{equation}
where $\omega(t)$ is a time-dependent weighting function.
Taking the gradient with respect to $\theta$ yields
\begin{equation}
    \label{eq:grad_gen}
    \nabla_\theta\mathcal{L}_{\text{DMD}}(\theta) = \int_t \omega(t) \bbE_{\bfx_0\sim p_{\theta,0},\bfx_t \sim q(\bfx_t|\bfx_0)}\left[\bfs_\phi(\bfx_t) - \bfs_\psi(\bfx_t) \right] \frac{\partial \bfx_t}{\partial \theta} \rmd t,
\end{equation}
where $\bfs_\psi(\bfx_t) \approx \nabla_{\bfx_t} \log q_t(\bfx_t)$ is the \textit{real score} provided by the frozen real model, and $\bfs_\phi(\bfx_t) \approx \nabla_{\bfx_t} \log p_{\theta,t}(\bfx_t)$ is the \textit{fake score} estimated by a trainable fake model. The fake model is optimized via denoising score matching~\citep{vincent2011connection}:
\begin{equation}
    \label{eq:loss_fake}
    \mathcal{L}_{\text{fake}}(\phi) = \int_t \lambda(t) \bbE_{\bfx_0\sim p_{\theta,0},\bfx_t \sim q(\bfx_t|\bfx_0)}\left[ \lVert \bfs_\phi(\bfx_t) - \nabla_{\bfx_t} \log q_t(\bfx_t|\bfx_0) \rVert_2^2 \right] \rmd t,
\end{equation}
where $\lambda(t)$ is a weighting function. DMD jointly optimizes the generator and the auxiliary fake model, but retains only $\mathbf{G}_\theta$ for few-step inference. See Appendix~\ref{app:related} for related work.

\subsection{Phase-Amplitude Decoupling}

The two-dimensional discrete Fourier transform (2D DFT), denoted by $\mathcal{F}$, transforms a spatial tensor $\bfx\in \mathbb{R}^{C \times H \times W}$ into a complex-valued spectrum $\mathcal{F}[\bfx] \in \mathbb{C}^{C \times H \times W}$. 
At each frequency coordinate $\bfu$, the spectrum can be expressed in polar form as
\begin{equation}
    \mathcal{F}[\bfx](\bfu) = \mathcal{A}_{\bfx}(\bfu) \cdot e^{i \mathcal{P}_{\bfx}(\bfu)},
\end{equation}
where $\mathcal{A}_{\bfx}(\bfu) \triangleq \left| \mathcal{F}[\bfx](\bfu) \right| \in \mathbb{R}_{\ge 0}$ denotes the amplitude of each frequency component, and $\mathcal{P}_{\bfx}(\bfu) \triangleq \angle\mathcal{F}[\bfx](\bfu) \in [-\pi, \pi]$ denotes its phase, which encodes the relative spatial arrangement.
The original spatial tensor $\bfx$ can be exactly reconstructed via the inverse DFT $\mathcal{F}^{-1}$:
\begin{equation}
    \bfx = \mathcal{F}^{-1} \left[ \mathcal{A}_{\bfx} \cdot e^{i \mathcal{P}_{\bfx}} \right].
\end{equation}
Compared with the traditional decomposition into real and imaginary parts, the polar form explicitly separates amplitude from phase.
Since modifying either the real or imaginary component generally changes both amplitude and phase, this representation is better suited to modulating spectral amplitudes while preserving phase.

\section{Method}

\subsection{Frequency-Domain Diagnostics of DMD}
\label{subsec:diagnostics}
To better understand the optimization dynamics of DMD, we analyze its generator update from a frequency-domain perspective. For clarity and visualization, we rewrite \eqref{eq:grad_gen} from score prediction to $\bfx_0$ prediction using \eqref{eq:parameterizations}:
\begin{equation}
    \label{eq:grad_gen_x0}
    \nabla_\theta\mathcal{L}_{\text{DMD}}(\theta) = \int_t \tilde{\omega}(t) \bbE_{\bfx_0\sim p_{\theta,0},\bfx_t \sim q(\bfx_t|\bfx_0)}\left[\bff_\phi(\bfx_t) - \bff_\psi(\bfx_t) \right] \frac{\partial \bfx_t}{\partial \theta} \rmd t,
\end{equation}
where $\tilde{\omega}(t)=\frac{\alpha_t}{\sigma_t^2} \omega(t)$ and the instantaneous directional error is defined as $\mathbf{\Delta}_t = \bff_\phi(\bfx_t) - \bff_\psi(\bfx_t)$.
In practice, this gradient can be implemented through the stop-gradient surrogate~\citep{yin2024improved}:
\begin{equation}
    \label{eq:dmd-pseudo}
    \mathcal{L}_{\text{DMD-pseudo}}(\theta) = \frac{1}{2}\int_t \tilde{\omega}(t) \bbE_{\bfx_0\sim p_{\theta,0},\bfx_t \sim q(\bfx_t|\bfx_0)} \left[ \| \bfx_t - \text{sg}\left[ \bfx_t - \mathbf{\Delta}_t \right] \|_2^2 \right] \rmd t.
\end{equation}
Although \eqref{eq:dmd-pseudo} does not define an independent global objective, its gradient at the current iterate recovers \eqref{eq:grad_gen_x0}. This surrogate gives an intuitive view of the generator update: $\bfx_t$ is locally driven toward a dynamic target determined by $\mathbf{\Delta}_t$, which then propagates to the generator through $\partial\bfx_t/\partial\theta$.

\begin{figure}[t]
    \centering
    \begin{subfigure}[b]{0.33\textwidth}
        \centering
        \includegraphics[width=\textwidth]{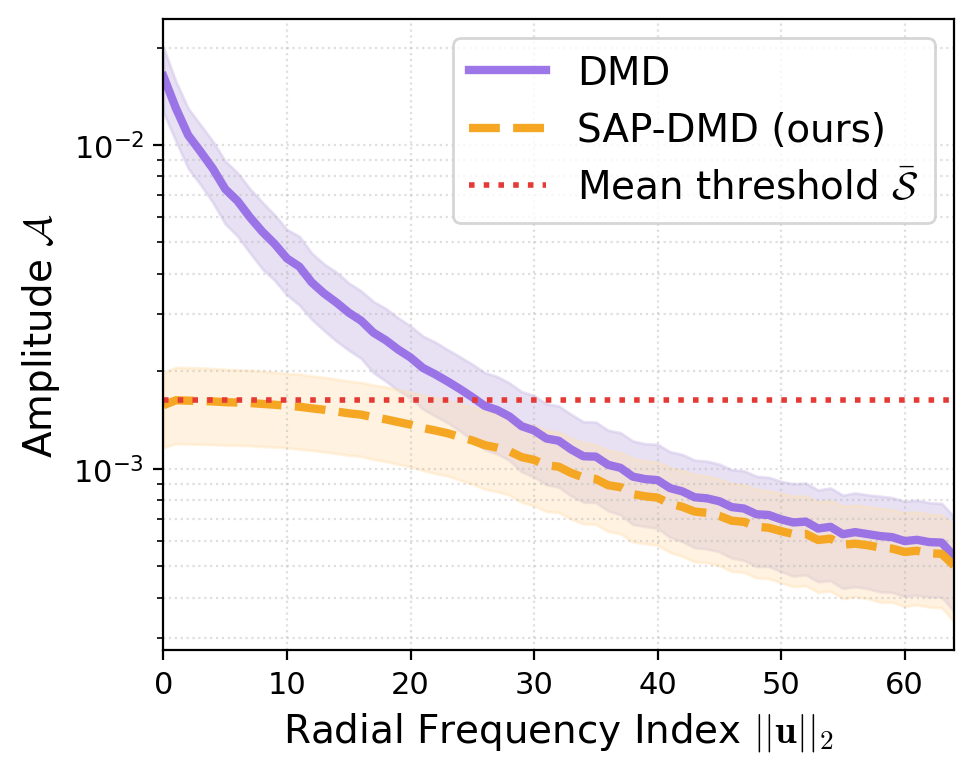}
        \caption{Radial amplitude suppression}
        \label{fig:diag_raps}
    \end{subfigure}
    \hfill
    \begin{subfigure}[b]{0.315\textwidth}
        \centering
        \includegraphics[width=\textwidth]{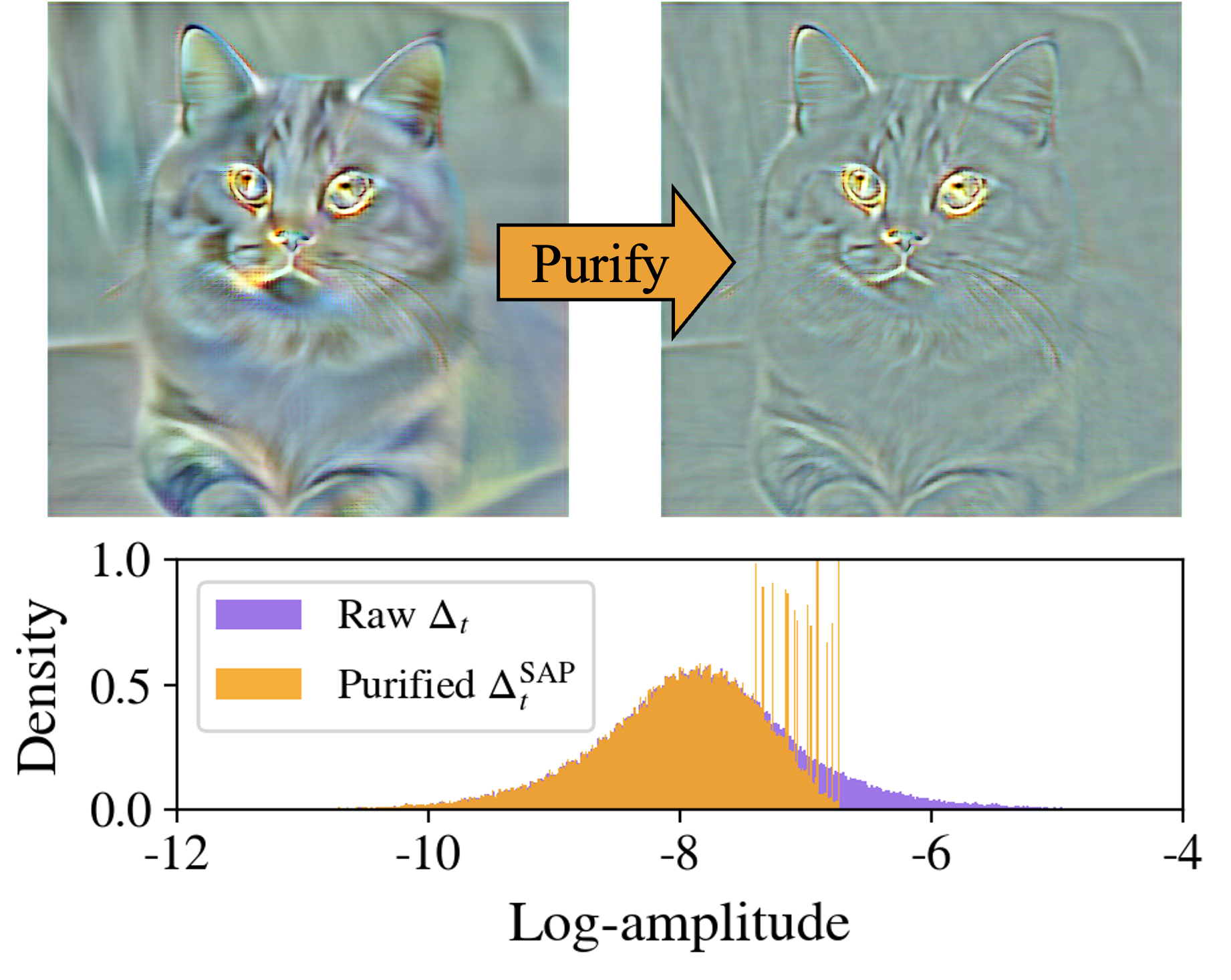}
        \caption{Amplitude distribution}
        \label{fig:purify}
    \end{subfigure}
    \hfill
    \begin{subfigure}[b]{0.335\textwidth}
        \centering
        \includegraphics[width=\textwidth]{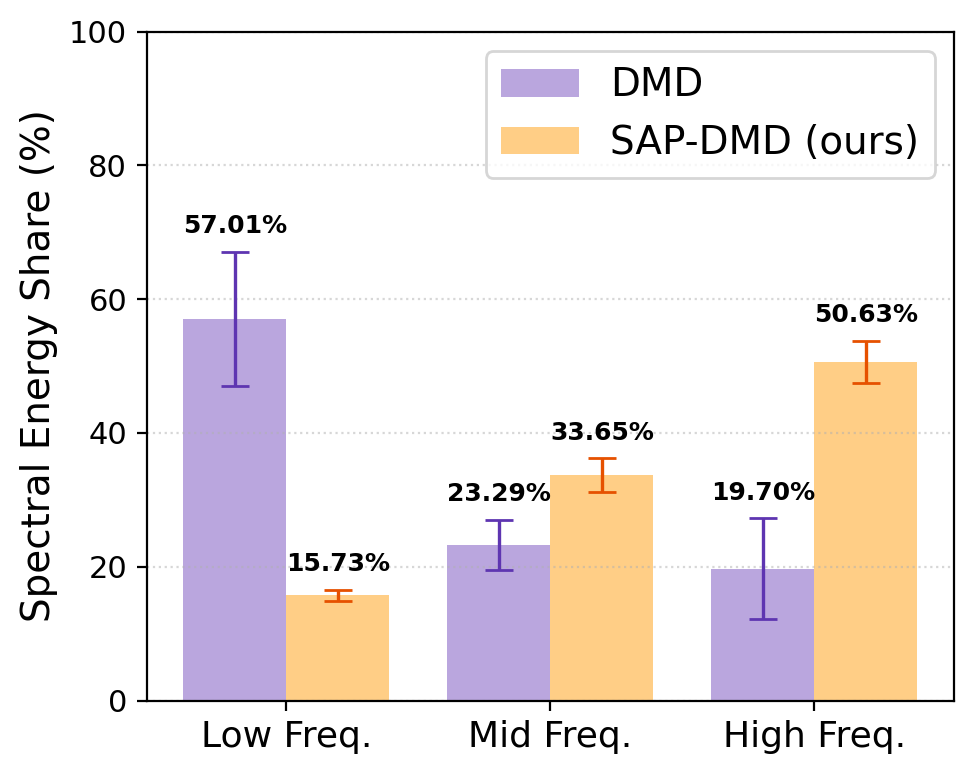}
        \caption{Spectral energy share}
        \label{fig:diag_energy}
    \end{subfigure}
    \caption{
    Frequency-domain diagnostics of the DMD directional error $\mathbf{\Delta}_t$ on SD3.5 Medium. Statistics are computed at $t=0.5$ from a DMD2 checkpoint trained for 200 steps and averaged over 1,000 samples.
    (a) Radial amplitude profiles before and after purification; the dotted line denotes the mean channel-wise threshold $\bar{\mathcal{S}}$.
    (b) Spatial visualization of $\mathbf{\Delta}_t$ before and after purification (top), with the corresponding log-amplitude distributions and channel-wise thresholds $\mathcal{S}_c$ (bottom).
    (c) Relative spectral energy shares of low-, mid-, and high-frequency components, defined by $\|\mathbf u\|_2<16$, $16\le\|\mathbf u\|_2<32$, and $32\le\|\mathbf u\|_2\le64$, respectively.
    }
    \label{fig:spectral_diagnostics_perfect_flow}
\end{figure}

The spatial formulation, however, does not explicitly reveal how different spatial scales contribute to this directional field. Fine textures, sharp boundaries, and coarse structures exhibit distinct spectral characteristics, making the frequency domain a natural space for examining their relative contributions. We therefore decompose $\mathbf{\Delta}_t$ using the 2D DFT:
\begin{equation}
    \mathcal{F}[\mathbf{\Delta}_t](\bfu) = \mathcal{A}_{\mathbf{\Delta}_t}(\bfu) \cdot e^{i \mathcal{P}_{\mathbf{\Delta}_t}(\bfu)}.
\end{equation}
Our analysis reveals a pronounced concentration of spectral amplitudes in the DMD directional field. As shown in Figure~\ref{fig:diag_raps}, the mean radial amplitude decays rapidly toward higher frequencies.
The purified profile shows that \ourName selectively suppresses the largest spectral amplitudes while leaving weaker components largely unchanged.
The log-amplitude distribution in Figure~\ref{fig:purify} further reveals a pronounced upper tail, indicating a small number of dominant spectral components.
Combined with Figure~\ref{fig:diag_raps}, this shows that these components are concentrated mainly at low frequencies, giving coarse-scale discrepancies a disproportionately large influence on the directional signal.

The spectral energy statistics in Figure~\ref{fig:diag_energy} provide a complementary view. 
Before purification, low-frequency components account for the majority of the spectral energy of $\mathbf{\Delta}_t$, 
while \ourName substantially reduces this concentration and increases the relative contributions of mid- and high-frequency components. 
Such spectral concentration may bias the generator update toward coarse-scale corrections and under-emphasize weaker fine-scale signals.
This behavior is consistent with the delayed recovery of fine-grained details observed during DMD training in Figure~\ref{fig:teaser}. 
Together, these observations motivate us to regulate the amplitude spectrum of $\mathbf{\Delta}_t$ by suppressing dominant components while preserving informative weaker signals.

\renewcommand{\algorithmiccomment}[1]{\hfill $\triangleright$ #1}
\begin{algorithm}[t]
\caption{Spectral Amplitude Purification for Distribution Matching Distillation}
\label{alg:alg}
\begin{algorithmic}[1]
\REQUIRE real model $\bff_\psi$, learning rates $\eta_f$ and $\eta_g$, scaling kernel $\kappa(\cdot,\cdot)$.
\ENSURE optimized generator $\bfG_\theta$.
\STATE Initialize $\bfG_\theta$ and the fake model $\bff_\phi$ from $\bff_\psi$.
\WHILE{not converged}
    \STATE Sample $t$, $\bfx_0 \sim p_{\theta,0}$, and $\bfx_t \sim q(\bfx_t|\bfx_0)$.
    \STATE $\phi \leftarrow \phi - \eta_f \nabla_\phi \mathcal{L}_{\text{fake}}(\phi)$
    \COMMENT{Update the fake model via \eqref{eq:loss_fake}}
    \STATE $\mathbf{\Delta}_t \leftarrow \bff_\phi(\bfx_t) - \bff_\psi(\bfx_t)$
    \COMMENT{Compute the directional error}
    \STATE $\mathbf{\Delta}_t^{\text{SAP}} \leftarrow \mathcal{T}[\mathbf{\Delta}_t]$
    \COMMENT{Purify via \eqref{eq:purify}}
    \STATE $\theta \leftarrow \theta - \eta_g
    \nabla_\theta \mathcal{L}_{\text{SAP-pseudo}}(\theta)$
    \COMMENT{Update the generator via \eqref{eq:sap_pseudo}}
\ENDWHILE
\end{algorithmic}
\end{algorithm}

\subsection{Spectral Amplitude Purification}
We introduce Spectral Amplitude Purification (SAP), a plug-and-play mechanism designed to mitigate the dominance of large spectral amplitudes identified in the DMD directional field. 
Rather than directly propagating the raw directional error $\mathbf{\Delta}_t$, SAP modulates its amplitude spectrum to suppress dominant components while preserving weaker spectral signals.

Formally, we define the purified directional error as
\begin{equation}
    \label{eq:purify}
    \mathbf{\Delta}_{t}^{\text{SAP}} = \mathcal{T}[\mathbf{\Delta}_t] = \mathcal{F}^{-1} \Big[ \kappa \big(\bfu, \mathcal{A}_{\mathbf{\Delta}_t}(\bfu)\big) \cdot \mathcal{F}[\mathbf{\Delta}_t](\bfu) \Big],
\end{equation}
where $\Omega$ denotes the set of frequency coordinates and $\kappa: \Omega \times \mathbb{R}_{\ge 0} \to \mathbb{R}_{\ge 0}$ is a real-valued, nonlinear scaling kernel. Since $\kappa$ is real and non-negative, SAP modifies only the spectral amplitude while leaving the phase unchanged:
\begin{equation} 
    \mathcal{F}[\mathbf{\Delta}^{\text{SAP}}_t](\bfu) = \underbrace{ \kappa \big( \bfu, \mathcal{A}_{\mathbf{\Delta}_t}(\bfu) \big) \mathcal{A}_{\mathbf{\Delta}_t}(\bfu) }_{\text{purified amplitude}} e^{i\mathcal{P}_{\mathbf{\Delta}_t}(\bfu)}. 
\end{equation}
This amplitude-only modulation directly targets the spectral concentration observed in Section~\ref{subsec:diagnostics} without introducing additional phase perturbations.

The scaling kernel is designed according to two principles: 
(1) \textbf{Small-Amplitude Preservation}, where weak spectral components should remain nearly unchanged; and
(2) \textbf{Large-Amplitude Suppression}, where unusually large components should be compressed.
As observed in Figure~\ref{fig:spectral_diagnostics_perfect_flow} and quantitatively validated in Appendix~\ref{app:diagnostics}, the raw spectral amplitudes are highly skewed, whereas the log transform yields a substantially more symmetric, bell-shaped distribution.
We therefore define an adaptive boundary using a one-sided $k$-sigma criterion in log-amplitude space.
For each sample and channel $c$, we estimate
\begin{equation}
    \mu_{c} = \mathbb{E}_{\bfu}[\log(\mathcal{A}_{\mathbf{\Delta}_t, c}(\bfu) + \epsilon)],
    \qquad
    s_{c} = \sqrt{\mathbb{E}_{\bfu}[(\log(\mathcal{A}_{\mathbf{\Delta}_t, c}(\bfu) + \epsilon) - \mu_c)^2]},
\end{equation}
where $\epsilon = 10^{-6}$ ensures numerical stability. 
We define the adaptive channel-wise threshold as $\mathcal{S}_c = \exp(\mu_{c} + k \cdot s_{c})$, mapping the log-domain $k$-sigma boundary back to the original amplitude domain. This threshold provides a sample- and channel-adaptive criterion for identifying dominant upper-tail amplitudes.
Given the adaptive boundary, the scaling kernel is defined as
\begin{equation}
    \label{eq:kernel}
    \kappa_c(\bfu, \mathcal{A}_{\mathbf{\Delta}_t, c}(\bfu)) = \left( 1 + \left( \frac{\mathcal{A}_{\mathbf{\Delta}_t, c}(\bfu)}{\mathcal{S}_c} \right)^p \right)^{-\frac{1}{p}},
\end{equation}
where $p$ controls the shape of the modulation. 
This kernel provides a smooth approximation to clipping the amplitude at $\mathcal{S}_c$: for $\mathcal{A}_{\mathbf{\Delta}_t,c}(\mathbf{u})\ll \mathcal{S}_c$, $\kappa_c\approx1$, leaving weak components nearly unchanged; for amplitudes substantially larger than $\mathcal{S}_c$, the purified amplitude approaches $\mathcal{S}_c$.
As $p\rightarrow\infty$, the modulation converges to hard clipping,
$\kappa_c\!\left(\mathbf{u},\mathcal{A}_{\mathbf{\Delta}_t,c}(\mathbf{u})\right) \mathcal{A}_{\mathbf{\Delta}_t,c}(\mathbf{u}) \rightarrow \min\!\left( \mathcal{A}_{\mathbf{\Delta}_t,c}(\mathbf{u}),\mathcal{S}_c \right)$,
while a smaller $p$ yields a smoother saturation.
As shown in Figure~\ref{fig:spectral_diagnostics_perfect_flow}, this adaptive modulation suppresses dominant spectral amplitudes while largely preserving weaker components.

SAP is incorporated into DMD by simply replacing the raw directional error $\mathbf{\Delta}_t$ with $\mathbf{\Delta}_t^{\text{SAP}}$. The resulting stop-gradient surrogate is
\begin{equation}
\label{eq:sap_pseudo}
    \mathcal{L}_{\text{SAP-pseudo}}(\theta) = \frac{1}{2} \int_t \tilde{\omega}(t) \mathbb{E}_{\bfx_0 \sim p_{\theta,0}, \bfx_t \sim q(\bfx_t|\bfx_0)} \left[ \left\| \bfx_t - \text{sg} \left[ \bfx_t - \mathbf{\Delta}_t^{\text{SAP}} \right] \right\|_2^2 \right] \rmd t.
\end{equation}
In this way, \ourName directly preconditions the DMD directional field without modifying the model architecture. By reducing the disproportionate influence of dominant spectral components, \ourName increases the relative contribution of weaker frequency components and accelerates the recovery of fine-grained structures, as shown in Figure~\ref{fig:evolve}. Algorithm~\ref{alg:alg} summarizes the training procedure. We next analyze how this preconditioning affects the distribution-matching dynamics.

\subsection{Theoretical Analysis}
\label{sec:theory}

We analyze SAP-DMD from the distribution-matching perspective.
For a fixed diffusion timestep $t$, we introduce an auxiliary time variable $\tau$
that describes the distributional evolution induced by \ourName, and denote
\begin{equation}
    \mathcal{E}_t(\tau) = \mathcal{D}_{\mathrm{KL}} \big(p_{\theta,t}(\tau)\|q_t\big).
\end{equation}
Under ideal score estimation, $\mathbf{\Delta}_t$ is proportional to the relative score $\nabla_{\mathbf{x}_t}\log(p_{\theta,t}(\tau)/q_t)$ up to a positive time-dependent factor, which we absorb into $\tilde{\omega}(t)$.

\begin{theorem}[KL Descent under the Idealized \ourName Flow] \label{thm:kl_descent} 
Assume $\tilde{\omega}(t)>0$ and $\kappa_c(\mathbf{u},\cdot)>0$ for all $c$ and $\mathbf{u}\in\Omega$. Under ideal score estimation and standard regularity conditions, 
\begin{equation} 
    \frac{\mathrm{d}}{\mathrm{d}\tau}\mathcal{E}_t = -\tilde{\omega}(t) \mathbb{E}_{p_{\theta,t}(\tau)} \left[ \sum_c\sum_{\mathbf{u}\in\Omega} \kappa_c\!\left( \mathbf{u}, \mathcal{A}_{\mathbf{\Delta}_t,c}(\mathbf{u}) \right) \mathcal{A}_{\mathbf{\Delta}_t,c}^2(\mathbf{u}) \right] \leq 0, 
\end{equation} 
with equality if and only if $p_{\theta,t}(\tau)=q_t$ almost everywhere. \end{theorem}

A complete proof is provided in Appendix~\ref{app:theory}.
Theorem~\ref{thm:kl_descent} characterizes \ourName as a positive spectral preconditioner of the idealized DMD distributional flow: it reweights the contribution of each frequency to KL dissipation without changing the distribution-matching fixed point.
For the kernel in \eqref{eq:kernel}, $0<\kappa_c\leq1$, so dominant spectral amplitudes are selectively suppressed while remaining positively aligned with the original directional field.
\section{Experiments}

\subsection{Experimental Setup}
\textbf{Models and baselines}. We conduct experiments on three representative text-to-image backbones: PixArt-$\alpha$~\citep{chen2024pixart}, Stable Diffusion 3 Medium, and Stable Diffusion 3.5 Medium~\citep{esser2024scaling}. We compare \ourName with a broad range of few-step distillation methods, including: (i) distribution matching methods, DMD2~\citep{yin2024improved}, Flash~\citep{chadebec2025flash}, TDM~\citep{luo2025learning}, SenseFlow~\citep{ge2025senseflow}, and Decoupled DMD~\citep{liu2025decoupled}; (ii) consistency-based methods, PCM~\citep{wang2024phased}, Hyper-SD~\citep{ren2024hyper}, and SANA-Sprint~\citep{chen2025sana}; and (iii) adversarial distillation methods, YOSO~\citep{luo2025you}, Turbo~\citep{sauer2024adversarial}, and SDXL-Lightning~\citep{lin2024sdxl}. We further include models from the FLUX.1 series~\citep{flux} to broaden the comparison under the 2-step setting.

\textbf{Training}. We train both 4-step and 2-step generators together with their auxiliary fake models using LoRA~\citep{hu2022lora} with rank $r=32$ and scaling factor $\alpha=32$. We use AdamW~\citep{loshchilov2017decoupled} with learning rates of $1\times10^{-5}$ for the generator and $5\times10^{-5}$ for the fake model. Unless otherwise specified, \ourName uses $p=\infty$, with $k=3$ for PixArt-$\alpha$ and $k=1$ for SD3 and SD3.5. Training consists of 4K iterations without GAN loss, followed by 2K additional iterations with GAN supervision. For a controlled comparison, we re-implement DMD2, TDM, SenseFlow, and Decoupled DMD using the same training recipe while retaining their respective method-specific components. All experiments are conducted on 4 NVIDIA A100 GPUs, with detailed configurations provided in Appendix~\ref{app:implementation}.

\textbf{Evaluation}. Sampling cost is measured by the number of function evaluations (NFE). We evaluate generation quality on 5K prompts from MS-COCO~\citep{lin2014microsoft} using ImageReward (IR)~\citep{xu2023imagereward}, PickScore (PS)~\citep{kirstain2023pick}, and CLIP Score (CS)~\citep{radford2021learning}. We further evaluate human-preference alignment on 3K HPSv2.1~\citep{wu2023human} prompts across its four categories: Anime, Concept-art, Painting, and Photo.

\begin{table}[t]
    \centering
    \fontsize{8pt}{9.2pt}\selectfont
    \caption{Quantitative results of 4-step generation on MS-COCO and HPSv2.1. $\dagger$: results from our re-implementation.
    $\times2$ indicates classifier-free guidance~\citep{ho2022classifier} requiring two NFEs per sampling step. GAN: with adversarial training enabled. Gray rows: multi-step base models. Best results are in bold and second best underlined.
    }
    \label{tab:comparison_4}
    \begin{tabularx}{\textwidth}{lcccccXXXXX}
        \toprule
        \multirow{2}{*}{\textbf{Method}} & \multirow{2}{*}{\textbf{NFE}} & \multirow{2}{*}{\textbf{GAN}} & \multicolumn{3}{c}{\textbf{MS-COCO}} & \multicolumn{5}{c}{\textbf{HPSv2.1}} \\
        \cmidrule(lr){4-6} \cmidrule(lr){7-11}
        & & & \textbf{IR} & \textbf{PS} & \textbf{CS} & \textbf{Anime} & \textbf{Concept} & \textbf{Painting} & \textbf{Photo} & \textbf{Avg.} \\
        \midrule
        \addlinespace[2pt]
        \multicolumn{10}{c}{\textbf{PixArt-$\alpha$} $(512 \times 512)$}  \\
        \addlinespace[-1pt]
        \midrule
        \rowcolor[gray]{0.9} PixArt-$\alpha$  & 25$\times$2 & \multicolumn{1}{c|}{} & 0.7893 & 22.4067 & \multicolumn{1}{c|}{30.47} & 30.87 & 29.04 & 28.69 & 28.43 & 29.26 \\
        YOSO                     & 4 & \multicolumn{1}{c|}{\checkmark}             & 0.7730 & 21.7805 & \multicolumn{1}{c|}{27.87} & 30.43 & 30.44 & 30.38 & 27.42 & 29.67 \\
        DMD2$\dagger$                     & 4 & \multicolumn{1}{c|}{}             & 0.7945 & 22.2972 & \multicolumn{1}{c|}{30.53} & 32.20 & 31.27 & 31.27 & 28.68 & 30.86 \\
        DMD2$\dagger$                     & 4 & \multicolumn{1}{c|}{\checkmark}             & 0.8047 & 22.4519 & \multicolumn{1}{c|}{30.82} & 32.35 & 31.33 & 31.35 & 29.03 & 31.02 \\
        TDM$\dagger$                      & 4 & \multicolumn{1}{c|}{}             & 0.8413 & 22.4650 & \multicolumn{1}{c|}{30.93} & 32.57 & 31.80 & 31.72 & 29.64 & 31.43 \\
        \textbf{SAP-DMD}  & 4 & \multicolumn{1}{c|}{}             & \underline{0.8761} & \underline{22.5569} & \multicolumn{1}{c|}{\underline{31.13}} & \underline{33.12} & \underline{32.11} & \underline{32.20} & \underline{29.83} & \underline{31.81} \\
        \textbf{SAP-DMD}  & 4 & \multicolumn{1}{c|}{\checkmark}             & \textbf{0.8805} & \textbf{22.6452} & \multicolumn{1}{c|}{\textbf{31.29}} & \textbf{33.23} & \textbf{32.30} & \textbf{32.39} & \textbf{29.99} & \textbf{31.98} \\
        
        \midrule
        \addlinespace[2pt]
        \multicolumn{10}{c}{\textbf{Stable Diffusion 3 Medium} $(1024 \times 1024)$}  \\
        \addlinespace[-1pt]
        \midrule
        \rowcolor[gray]{0.9} SD3 Medium  & 25$\times$2 & \multicolumn{1}{c|}{}  & 1.0026 & 22.5014 & \multicolumn{1}{c|}{32.19} & 30.63 & 29.91 & 30.29 & 27.62 & 29.61 \\
        Hyper-SD                & 4$\times$2 & \multicolumn{1}{c|}{}            & 0.8687 & 22.4059 & \multicolumn{1}{c|}{\textbf{31.33}} & 31.08 & 29.36 & 29.91 & 27.01 & 29.34 \\
        PCM                     & 4 & \multicolumn{1}{c|}{\checkmark}           & 0.6384 & 22.2444 & \multicolumn{1}{c|}{30.54} & 29.85 & 28.37 & 28.87 & 26.24 & 28.33 \\
        Flash                   & 4 & \multicolumn{1}{c|}{\checkmark}           & 0.9009 & 22.5143 & \multicolumn{1}{c|}{\underline{31.25}} & 30.32 & 29.40 & 30.11 & 27.06 & 29.22 \\
        DMD2$\dagger$                    & 4 & \multicolumn{1}{c|}{}                     & 0.9165 & 22.3188 & \multicolumn{1}{c|}{30.53} & 31.35 & 30.21 & 30.72 & 29.32 & 30.40 \\
        DMD2$\dagger$                    & 4 & \multicolumn{1}{c|}{\checkmark}           & \underline{1.0026} & \underline{22.6412} & \multicolumn{1}{c|}{30.77} & \underline{31.95} & \underline{30.97} & \underline{31.35} & \underline{29.33} & \underline{30.90} \\
        \textbf{SAP-DMD} & 4 & \multicolumn{1}{c|}{}                     & 0.9903 & 22.4440 & \multicolumn{1}{c|}{30.74} & 31.84 & 30.71 & 30.91 & \textbf{29.39} & 30.71 \\
        \textbf{SAP-DMD} & 4 & \multicolumn{1}{c|}{\checkmark}           & \textbf{1.0434} & \textbf{22.7133} & \multicolumn{1}{c|}{30.89} & \textbf{32.92} & \textbf{31.64} & \textbf{31.85} & 29.23 & \textbf{31.41} \\
        \midrule
        \addlinespace[2pt]
        \multicolumn{10}{c}{\textbf{Stable Diffusion 3.5 Medium} $(1024 \times 1024)$}  \\
        \addlinespace[-1pt]
        \midrule
        \rowcolor[gray]{0.9} SD3.5 Medium  & 25$\times$2 & \multicolumn{1}{c|}{} & 1.0208 & 22.4234 & \multicolumn{1}{c|}{32.78} & 31.32 & 30.49 & 30.48 & 27.66 & 29.99 \\
        Turbo                   & 4$\times$2 & \multicolumn{1}{c|}{\checkmark} & 0.5763 & 21.9061 & \multicolumn{1}{c|}{31.23} & 29.33 & 27.17 & 28.03 & 26.09 & 27.65 \\
        DMD2$\dagger$                    & 4 & \multicolumn{1}{c|}{} & 1.0960 & 22.5591 & \multicolumn{1}{c|}{31.44} & \underline{34.11} & \underline{33.46} & 33.66 & 30.45 & 32.92 \\
        DMD2$\dagger$                    & 4 & \multicolumn{1}{c|}{\checkmark} & \underline{1.1217} & \underline{22.7215} & \multicolumn{1}{c|}{31.63} & 33.68 & 32.99 & 33.26 & 30.10 & 32.51 \\
        TDM$\dagger$                     & 4 & \multicolumn{1}{c|}{} & 1.0689 & 22.5779 & \multicolumn{1}{c|}{\textbf{31.83}} & 33.04 & 32.72 & 32.82 & 29.15 & 31.93 \\
        SenseFlow$\dagger$               & 4 & \multicolumn{1}{c|}{} & 1.0792 & 22.5966 & \multicolumn{1}{c|}{\textbf{31.83}} & 33.93 & 33.10 & 33.44 & 29.46 & 32.48 \\
        Decoupled DMD$\dagger$           & 4 & \multicolumn{1}{c|}{} & 1.0422 & 22.5071 & \multicolumn{1}{c|}{31.55} & 33.57 & 32.42 & 32.95 & 29.50 & 32.11 \\
        \textbf{SAP-DMD} & 4 & \multicolumn{1}{c|}{} & 1.1120 & 22.5980 & \multicolumn{1}{c|}{\underline{31.81}} & \textbf{34.42} & \textbf{33.77} & \textbf{33.89} & \underline{30.61} & \textbf{33.17} \\
        \textbf{SAP-DMD} & 4 & \multicolumn{1}{c|}{\checkmark} & \textbf{1.1401} & \textbf{22.7480} & \multicolumn{1}{c|}{31.77} & 34.03 & 33.38 & \underline{33.72} & \textbf{30.69} & \underline{32.96} \\
        \bottomrule
    \end{tabularx}
\end{table}

\begin{table}[t]
    \centering
    \fontsize{8pt}{9.2pt}\selectfont
    \caption{Cross-model comparison of 2-step generation on MS-COCO and HPSv2.1. $\dagger$: results from our re-implementation. GAN: with adversarial training enabled. Gray rows: multi-step base models. Best results are in bold and second best underlined.}
    \label{tab:comparison_2}
    \begin{tabularx}{\textwidth}{lcccccXXXXX}
        \toprule
        \multirow{2}{*}{\textbf{Method}} & \multirow{2}{*}{\textbf{NFE}} & \multirow{2}{*}{\textbf{GAN}} & \multicolumn{3}{c}{\textbf{MS-COCO}} & \multicolumn{5}{c}{\textbf{HPSv2.1}} \\
        \cmidrule(lr){4-6} \cmidrule(lr){7-11}
        & & & \textbf{IR} & \textbf{PS} & \textbf{CS} & \textbf{Anime} & \textbf{Concept} & \textbf{Painting} & \textbf{Photo} & \textbf{Avg.} \\
        \midrule
        \rowcolor[gray]{0.9} SDXL           & 25$\times$2 & \multicolumn{1}{c|}{} & 0.7586 & 22.4153 & \multicolumn{1}{c|}{32.76} & 30.32 & 28.55 & 28.15 & 26.95 & 28.49 \\
        \rowcolor[gray]{0.9} SD3 Medium     & 25$\times$2 & \multicolumn{1}{c|}{} & 1.0026 & 22.5014 & \multicolumn{1}{c|}{32.19} & 30.63 & 29.91 & 30.29 & 27.62 & 29.61 \\
        \rowcolor[gray]{0.9} SD3.5 Medium   & 25$\times$2 & \multicolumn{1}{c|}{} & 1.0208 & 22.4234 & \multicolumn{1}{c|}{32.78} & 31.32 & 30.49 & 30.48 & 27.66 & 29.99 \\
        \rowcolor[gray]{0.9} FLUX.1-dev     & 25          & \multicolumn{1}{c|}{} & 1.0653 & 23.0421 & \multicolumn{1}{c|}{31.50} & 32.08 & 30.38 & 30.88 & 29.40 & 30.68 \\
        SDXL-Lightning          & 2 & \multicolumn{1}{c|}{\checkmark} & 0.6511 & 22.3895 & \multicolumn{1}{c|}{31.28} & 30.86 & 29.53 & 29.49 & 27.23 & 29.28 \\
        PCM-SD3                 & 2 & \multicolumn{1}{c|}{\checkmark} & 0.5722 & 21.8711 & \multicolumn{1}{c|}{30.60} & 28.94 & 27.78 & 28.78 & 26.16 & 27.92 \\
        SANA-Sprint             & 2 & \multicolumn{1}{c|}{\checkmark} & 1.0168 & \textbf{22.8824} & \multicolumn{1}{c|}{31.71} & 31.88 & 30.05 & 29.83 & 29.32 & 30.27 \\
        FLUX.1-schnell          & 2 & \multicolumn{1}{c|}{\checkmark} & 1.0186 & \underline{22.6435} & \multicolumn{1}{c|}{\textbf{32.30}} & 30.92 & 29.71 & 29.91 & 28.47 & 29.75 \\
        DMD2-SD35$\dagger$               & 2 & \multicolumn{1}{c|}{} & 0.9623 & 22.1067 & \multicolumn{1}{c|}{31.57} & 32.97 & 31.74 & 31.94 & 28.43 & 31.27 \\
        DMD2-SD35$\dagger$               & 2 & \multicolumn{1}{c|}{\checkmark} & 1.0465 & 22.3839 & \multicolumn{1}{c|}{31.87} & 32.77 & 31.48 & 31.83 & 28.46 & 31.14 \\
        \textbf{SAP-DMD-SD35}   & 2 & \multicolumn{1}{c|}{} & \underline{1.0867} & 22.4414 & \multicolumn{1}{c|}{31.87} & \textbf{34.16} & \textbf{33.17} & \textbf{33.35} & \textbf{29.99} & \textbf{32.67} \\
        \textbf{SAP-DMD-SD35}   & 2 & \multicolumn{1}{c|}{\checkmark} & \textbf{1.1104} & 22.5945 & \multicolumn{1}{c|}{\underline{32.14}} & \underline{33.58} & \underline{32.92} & \underline{33.04} & \underline{29.97} & \underline{32.38} \\
        \bottomrule
    \end{tabularx}
\end{table}

\subsection{Main Results}
We evaluate \ourName quantitatively and qualitatively in the extreme $2$-step and $4$-step regimes.

\begin{figure}[t]
    \centering
    \includegraphics[width=0.96\textwidth]{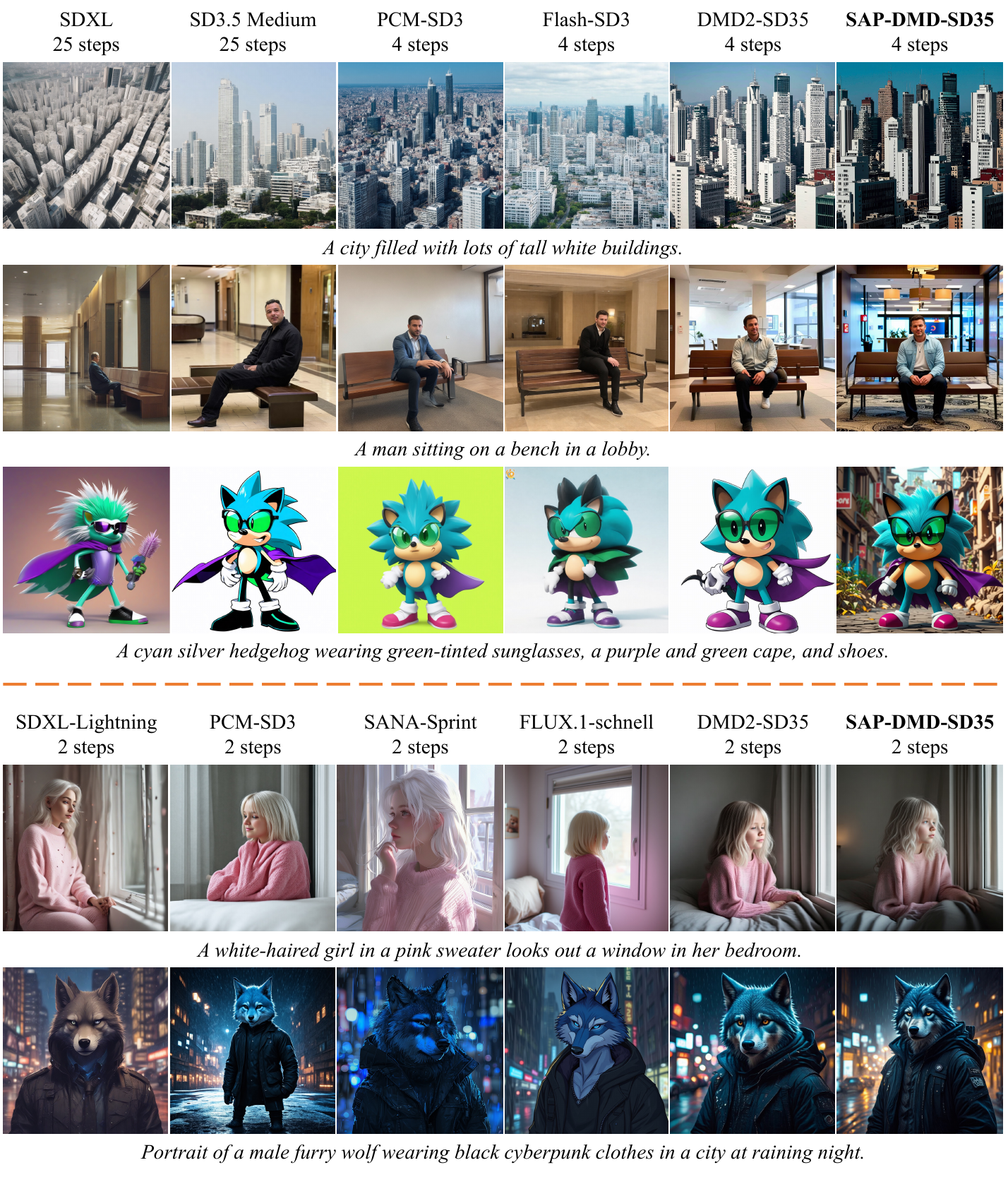}
    \caption{Qualitative comparison of 4-step (top) and 2-step (bottom) generation.}
    \label{fig:qualitative}
\end{figure}

\textbf{4-step generation}.
As shown in Table~\ref{tab:comparison_4}, \ourName consistently improves over DMD2 across PixArt-$\alpha$, SD3, and SD3.5, with particularly strong gains in ImageReward and HPSv2.1.
Without GAN supervision, \ourName already matches or surpasses competitive distribution-matching baselines, while the GAN-enhanced variant further achieves the best overall preference scores on SD3.
On SD3.5, \ourName also outperforms TDM, SenseFlow, and Decoupled DMD on the main preference metrics.
Qualitative results in Figure~\ref{fig:qualitative} further demonstrate improved fine-detail quality with only four inference steps.

\textbf{2-step generation}.
As shown in Table~\ref{tab:comparison_2}, \ourName remains highly competitive under the challenging 2-step setting.
Without GAN supervision, \ourName already achieves the best HPSv2.1 average score among all compared 2-step methods, while substantially outperforming DMD2 on ImageReward, CLIP Score, and HPSv2.1.
With GAN supervision, \ourName further achieves the highest ImageReward among the evaluated 2-step models.
Figure~\ref{fig:qualitative} also shows that \ourName preserves sharp structures and fine visual details with only two inference steps.

\textbf{Training acceleration}.
SAP-DMD introduces negligible computational overhead. Under identical hardware and training settings, DMD2 and \ourName require 10.53 and 10.56 A100 GPU hours, respectively.
Meanwhile, as shown in Figure~\ref{fig:evolve} and Appendix~\ref{sec:additional_detail_recovery}, \ourName substantially accelerates the early-stage recovery of fine-grained details compared with DMD2, accompanied by consistently faster improvements across multiple evaluation metrics.
This faster recovery enables \ourName to reach higher generation quality within fewer training iterations, demonstrating improved training efficiency.

\begin{table*}[t]
    \centering
    \begin{minipage}[t]{0.47\textwidth}
        \centering
        \fontsize{8.5pt}{9pt}\selectfont
        \captionof{table}{Ablation study on the kernel shape parameter $p$ and the statistical boundary factor $k$ on SD3.5 Medium. Gray row: DMD2.}
        \label{tab:ablation}
        \begin{tabular}{cccccc}
            \toprule
            \multirow{2}{*}{\textbf{$p$}} &
            \multirow{2}{*}{\textbf{$k$}} &
            \multicolumn{3}{c}{\textbf{MS-COCO 5K}} &
            \multicolumn{1}{c}{\textbf{HPSv2.1}} \\
            \cmidrule(lr){3-5} \cmidrule(lr){6-6}
            & & \textbf{IR} & \textbf{PS} & \textbf{CS} & \textbf{Avg.} \\
            \midrule
            \rowcolor[gray]{0.9}
            - & - & 1.0960 & 22.5591 & 31.44 & 32.92 \\
            $\infty$ & 1 & 1.1120 & 22.5980 & \textbf{31.81} & 33.17 \\
            $\infty$ & 2 & 1.1195 & 22.6227 & 31.66 & \underline{33.43} \\
            $\infty$ & 3 & 1.1131 & 22.6160 & 31.71 & 33.21 \\
            3 & 1 & 1.0973 & 22.5869 & 31.61 & 33.41 \\
            3 & 2 & 1.1043 & 22.5819 & 31.66 & 33.39 \\
            3 & 3 & \textbf{1.1318} & \textbf{22.6562} & 31.66 & \textbf{33.51} \\
            1 & 1 & 1.1254 & 22.6230 & 31.70 & 33.42 \\
            1 & 2 & \underline{1.1284} & \underline{22.6482} & \underline{31.75} & 33.32 \\
            1 & 3 & 1.1260 & 22.6050 & 31.70 & 33.36 \\
            \bottomrule
        \end{tabular}
    \end{minipage}
    \hfill
    \begin{minipage}[t]{0.47\textwidth}
    \centering
    \fontsize{8.5pt}{9pt}\selectfont
    \captionof{table}{
        Design ablation on SD3.5 Medium.
        \textit{Phase}: phase-only modulation;
        \textit{S-Clip}: spatial-domain clipping;
        \textit{R-Mask}: random masking;
        \textit{LF-Sup}: fixed low-frequency suppression;
        \textit{SAP-DMD}: amplitude-only purification.
    }
    \label{tab:design_ablation}
    \setlength{\tabcolsep}{4pt}
    \begin{tabular}{lcccc}
        \toprule
        \multirow{2}{*}{\textbf{Variant}}
        & \multicolumn{3}{c}{\textbf{MS-COCO 5K}}
        & \multicolumn{1}{c}{\textbf{HPSv2.1}} \\
        \cmidrule(lr){2-4} \cmidrule(lr){5-5}
        & \textbf{IR} & \textbf{PS} & \textbf{CS} & \textbf{Avg.} \\
        \midrule
        \rowcolor[gray]{0.9}
        DMD2
            & 1.0960 & 22.5591 & 31.44 & 32.92 \\
        Phase
            & 1.0912 & 22.5630 & 31.49 & 32.68 \\
        S-Clip
            & 1.0383 & 22.5444 & 31.71 & 32.04 \\
        R-Mask
            & 1.0717 & 22.5085 & 31.45 & 32.30 \\
        LF-Sup
            & 1.0974 & 22.5696 & 31.56 & 32.65 \\
        \textbf{SAP-DMD}
            & \textbf{1.1120}
            & \textbf{22.5980}
            & \textbf{31.81}
            & \textbf{33.17} \\
        \bottomrule
    \end{tabular}
    \end{minipage}
\end{table*}

\begin{figure}[t]
    \centering
    \begin{subfigure}[b]{0.49\textwidth}
        \centering
        \includegraphics[width=\textwidth]{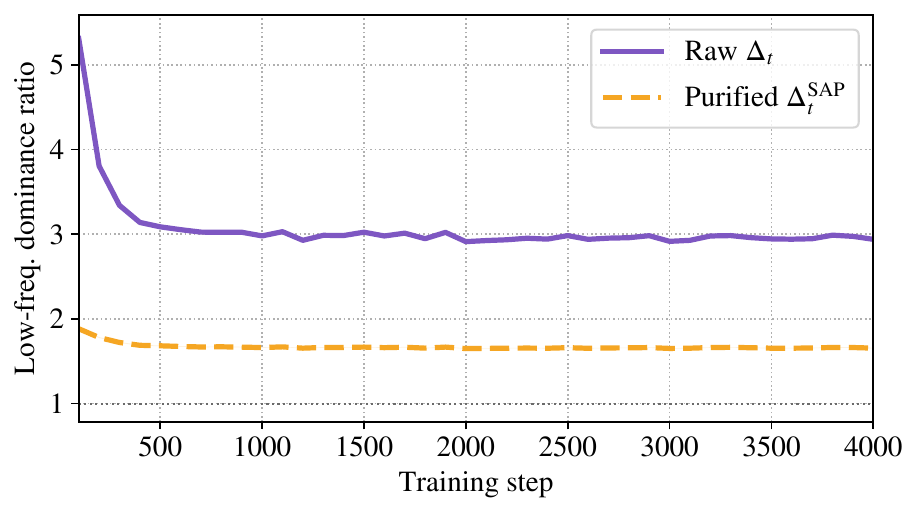}
        \caption{Low-frequency dominance evolution}
        \label{fig:dominance}
    \end{subfigure}
    \hfill
    \begin{subfigure}[b]{0.49\textwidth}
        \centering
        \includegraphics[width=\textwidth]{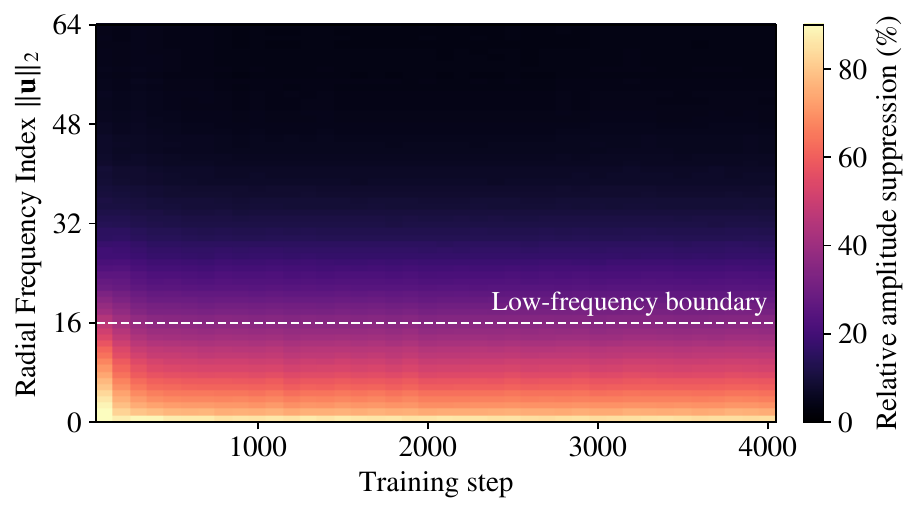}
        \caption{Frequency-wise suppression}
        \label{fig:suppression}
    \end{subfigure}
    \hfill
    \caption{
    Spectral evolution throughout SAP-DMD training on SD3.5 Medium. Statistics are averaged over 256 samples at $t=0.5$ with a checkpoint interval of 100 steps.
    (a) Low-frequency dominance $R_{\mathrm{LF}}$, defined as the ratio of mean amplitudes over $\|\mathbf{u}\|_2<16$ and $16\leq\|\mathbf{u}\|_2\leq64$, remains consistently reduced after purification.
    (b) Relative amplitude suppression, measured as $1-\bar{\mathcal{A}}^{\mathrm{SAP}}(\|\mathbf{u}\|_2)/\bar{\mathcal{A}}(\|\mathbf{u}\|_2)$, where $\bar{\mathcal{A}}$ denotes the sample-averaged radial amplitude.
    }
    \label{fig:spectral_evolution}
\end{figure}

\textbf{Ablation study}.
We study both the hyperparameter sensitivity and the design choices of \ourName on SD3.5.
As shown in Table~\ref{tab:ablation}, all tested \ourName configurations outperform the DMD2 baseline across the evaluated metrics, indicating robustness to the choice of $p$ and $k$.
While $p=3,k=3$ achieves the best preference-oriented scores, we adopt $p=\infty,k=1$ as the default setting since it yields the highest CLIP Score while remaining competitive on the other metrics.
Table~\ref{tab:design_ablation} further shows that phase-only modulation, spatial-domain clipping, random masking, and fixed low-frequency suppression fail to provide consistent improvements over DMD2, whereas amplitude purification improves all evaluated metrics.
These results indicate that the gains of \ourName stem from adaptive amplitude modulation rather than from generic gradient suppression or frequency-specific attenuation. Implementation details are provided in Appendix~\ref{app:implementation}.

\textbf{Spectral evolution}.
Figure~\ref{fig:spectral_evolution} further shows that the diagnosed spectral concentration persists throughout training.
\ourName consistently reduces the low-frequency dominance ratio (Figure~\ref{fig:dominance}), while the frequency-resolved suppression map (Figure~\ref{fig:suppression}) reveals substantially stronger attenuation at low frequencies.
These results demonstrate that \ourName provides persistent, adaptive spectral preconditioning beyond the early-stage snapshot in Figure~\ref{fig:diag_energy}.

\section{Conclusion}

In this paper, we provide a frequency-domain perspective on the optimization dynamics of DMD. Our analysis reveals a pronounced spectral amplitude concentration in the DMD directional field, where dominant low-frequency components overwhelm weaker mid- and high-frequency signals and delay fine-detail recovery. To address this issue, we introduce \ourName, a plug-and-play spectral preconditioning method that adaptively suppresses dominant amplitudes while preserving phase and weaker spectral components. 
Our idealized analysis further shows that SAP-DMD preserves the distribution-matching fixed point and maintains KL descent under ideal score estimation.
Experiments across multiple text-to-image backbones demonstrate faster convergence and improved 2-step and 4-step generation quality.

Despite these improvements, preserving complex semantic details remains challenging under limited sampling budgets. We hope our findings motivate further study of spectral distillation dynamics and frequency-aware optimization for more accurate and efficient few-step generation.

\section*{AI Use Statement}

Generative AI tools were used for language editing, feedback on experimental design, code refinement, implementation of several ablation experiments, and assistance with the theoretical analysis and proof writing. All AI-assisted code, mathematical derivations, and manuscript content were reviewed and verified by the authors. The authors take full responsibility for the final content and claims of this work.


\bibliography{main}
\bibliographystyle{iclr2027_conference}

\clearpage
\appendix
\setcounter{theorem}{0}

\section{Related Work}
\label{app:related}
\textbf{Few-step diffusion distillation}.
Diffusion distillation aims to reduce the number of sampling steps while maintaining high generation quality. Existing approaches can be broadly grouped into three major categories: trajectory distillation, consistency distillation, and distribution matching. 
\textbf{Trajectory distillation} methods~\citep{luhman2021knowledge,salimans2022progressive,meng2023distillation,zhou2024simple} align the intermediate states along the teacher's sampling trajectory with those of the student model, emphasizing temporal consistency and often relying on carefully designed schedules. 
\textbf{Consistency distillation}~\citep{song2023consistency,song2023improved,kim2023consistency,luo2023latent,wang2024phased,ren2024hyper,chen2025sana,geng2026mean} enforces output consistency across noise levels, simplifying training and improving sample quality, but typically requires careful hyperparameter tuning for stable optimization.
\textbf{Distribution Matching Distillation} (DMD) formulates few-step distillation as explicit distribution alignment~\citep{luo2023diff,yin2023one,nguyen2023swiftbrush, yin2024improved}. By minimizing the KL divergence between the student and teacher marginal distributions at each diffusion step, DMD provides an effective framework for training few-step generators to approximate the teacher distribution. 
Several extensions further improve scalability and generation fidelity. 
TDM~\citep{luo2025learning} partitions the optimization into non-overlapping intervals and introduces importance sampling for efficient fake model training.
SenseFlow~\citep{ge2025senseflow} introduces parameter shifting to stabilize fake model training and intra-segment guidance as additional supervision.
Decoupled DMD~\citep{liu2025decoupled} separates distribution matching from classifier-free guidance (CFG) augmentation through independent re-noising strategies. 
Different from these approaches, \ourName investigates the spectral characteristics of the DMD directional field and introduces a spectral amplitude modulation mechanism. By reducing the dominance of concentrated low-frequency components, \ourName improves few-step distillation without modifying the model architecture or training objective.

\textbf{Frequency-domain perspectives on diffusion models}.
Recent studies have revealed that diffusion models exhibit frequency-dependent generation dynamics. 
Fourier-space analysis shows that different frequency components undergo distinct corruption and recovery throughout the diffusion process, leading to a coarse-to-fine generation behavior~\citep{falck2025fourier}. 
Beyond analysis, recent methods incorporate frequency information into diffusion generation. 
FreqFlow~\citep{ren2026frequency} incorporates frequency-aware conditioning into flow matching by explicitly modeling low- and high-frequency components to improve generation quality, while FreeU~\citep{si2024freeu} and DeCo~\citep{ma2026deco} introduce frequency-aware modulation in intermediate representations and generation architectures, respectively. 
Frequency-aware spectral modeling has also been explored in diffusion compression~\citep{yang2023diffusion}. 
However, existing frequency-domain approaches mainly focus on generation dynamics, model representations, or reconstruction objectives, while the spectral properties of optimization signals in diffusion distillation remain largely underexplored. 
In contrast, \ourName studies the Fourier structure of the DMD directional field and introduces spectral amplitude modulation during distillation.

\section{Implementation Details}
\label{app:implementation}

\textbf{General settings}. We optimize the generator and the fake model using LoRA~\citep{hu2022lora} with hyperparameters $\alpha=32$ and rank $r=32$, employing the AdamW~\citep{loshchilov2017decoupled} optimizer. 
LoRA adapters are applied to the attention projections $\{\texttt{to\_q}, \texttt{to\_k}, \texttt{to\_v}, \texttt{to\_out.0}\}$.
Training is conducted in two phases. 
The generator is first trained for 4,000 iterations without GAN loss, using only the prompts of the JourneyDB~\citep{sun2023journeydb} dataset. 
Subsequently, the GAN loss is introduced for an additional 2,000 iterations using the MJHQ-30K~\citep{li2024playground} dataset to further improve generation quality.
For the GAN loss, we adopt the implementation from SenseFlow~\citep{ge2025senseflow}, where the discriminator leverages pretrained DINOv2~\citep{oquab2023dinov2} and CLIP~\citep{radford2021learning} encoders to provide semantically rich and spatially aligned supervision.
Due to limited computational resources, a global batch size of 4 is used in all experiments.
For a controlled comparison, we keep the common training setup fixed across the re-implemented distribution-matching baselines.
In the following, we detail the hyperparameters specific to each re-implemented method.

\textbf{DMD2} introduces several improvements over vanilla DMD: the Two Time-scale Update Rule (TTUR), which performs more updates on the fake score model per generator update; backward simulation, which mitigates the training--inference mismatch; and a GAN loss that enhances generation quality. 
In our experiments, following TDM, we disable TTUR and instead use a larger learning rate of $5\times10^{-5}$ for the fake model, while setting the generator learning rate to $1\times10^{-5}$.
We set the GAN loss weight to 0.1 in all second-stage training.
For generator training, we follow the standard protocol: the real model employs classifier-free guidance (e.g., 7.0 in our experiments), while the fake model does not.

\textbf{\ourName}. 
\ourName strictly follows the above training settings of DMD2. For the proposed amplitude modulation mechanism, we set the kernel shape parameter $p=\infty$, corresponding to a hard clipping function, and set the statistical boundary factor to $k=3$ for PixArt-$\alpha$ and $k=1$ for both SD3 and SD3.5.

\textbf{TDM} introduces a re-noising strategy that segments optimization into non-overlapping intervals. It further introduces importance sampling for training the fake model and employs a Pseudo-Huber loss for generator training. In our re-implementation of TDM, we enable these modifications while keeping the other settings consistent with those used in DMD2.

\textbf{SenseFlow}. For SenseFlow, we follow the practical implementation by shifting the fake model parameters toward the generator using $\phi \leftarrow 0.98 \cdot \phi + 0.02 \cdot \theta$. We also enable the proposed intra-segment guidance with a weight of 0.5. All other settings remain consistent with those of DMD2.

\textbf{Decoupled DMD} separates generator training into two components: a distribution matching (DM) term and a classifier-free guidance augmentation (CA) term. It applies independent re-noising timesteps to the two terms, constraining the CA re-noising timestep while leaving the DM re-noising timestep unconstrained. All other settings are kept consistent with those of DMD2.

\textbf{Design ablations}.
For the design ablations in Table~\ref{tab:design_ablation}, all variants modify the same DMD directional error while keeping the remaining training settings unchanged.
For \textit{Phase}, we preserve the spectral amplitude and apply the same statistical clipping rule only to the phase magnitude, with $p=\infty$ and $k=1$; the channel-wise threshold is given by the mean plus $k$ standard deviations of the absolute phase and capped at $\pi$.
For \textit{S-Clip}, the same hard-clipping rule is applied directly to the spatial-domain magnitude with $p=\infty$ and $k=1$.
For \textit{R-Mask}, conjugate-symmetric frequency pairs are independently masked with probability $0.1$, preserving a real-valued inverse transform.
For \textit{LF-Sup}, Fourier coefficients with normalized radial frequency $r\leq0.25$ are uniformly scaled by $0.5$, while the remaining frequencies are left unchanged.

\section{Theoretical Proof}
\label{app:theory}

For a fixed diffusion timestep $t$, we omit the subscript $t$ when no ambiguity arises.
Let $p(\mathbf{x})$ denote the density evolving with the auxiliary parameter $\tau$, initialized from $p_{\theta,t}$, and let $q(\mathbf{x})=q_t(\mathbf{x})$ and
$\mathbf{\Delta}(\mathbf{x})=\mathbf{\Delta}_t(\mathbf{x})$. 
We assume that the involved densities are sufficiently smooth, positive on a common connected support, and decay properly at the boundary so that integration by parts is valid.

Under ideal score estimation, the fake and real models recover the scores of $p$ and $q$, respectively. From the score--prediction relation in \eqref{eq:parameterizations}, the directional error in \eqref{eq:grad_gen_x0} satisfies
\begin{equation}
    \label{eq:relative_score}
    \mathbf{\Delta}(\mathbf{x}) = a_t \nabla_{\mathbf{x}}\log\frac{p(\mathbf{x})}{q(\mathbf{x})},
\end{equation}
where $a_t>0$ is a time-dependent factor determined by the diffusion parameterization. 

We consider the idealized distributional flow induced by \ourName,
\begin{equation}
    \label{eq:sap_velocity}
    \frac{\mathrm{d}\mathbf{x}}{\mathrm{d}\tau} = \mathbf{v}(\mathbf{x}) = -\tilde{\omega}(t)\,\mathcal{T}[\mathbf{\Delta}](\mathbf{x}).
\end{equation}
The corresponding density evolves according to the continuity equation
\begin{equation}
    \label{eq:continuity}
    \frac{\partial p(\mathbf{x})}{\partial \tau} = -\nabla_{\mathbf{x}} \cdot \big( p(\mathbf{x})\mathbf{v}(\mathbf{x}) \big).
\end{equation}

Recall that
\begin{equation}
    \mathcal{E}_t = \mathcal{D}_{\mathrm{KL}}(p\|q) = \int p(\mathbf{x}) \log \frac{p(\mathbf{x})}{q(\mathbf{x})}\,\mathrm{d}\mathbf{x}.
\end{equation}
Differentiating with respect to $\tau$ gives
\begin{align}
    \label{eq:kl_derivative_1}
    \frac{\mathrm{d}}{\mathrm{d}\tau}\mathcal{E}_t
    & = \int \frac{\partial p}{\partial\tau} \left( \log\frac{p}{q}+1 \right) \mathrm{d}\mathbf{x} \nonumber \\
    & = \int \frac{\partial p}{\partial\tau} \log\frac{p}{q}\,\mathrm{d}\mathbf{x},
\end{align}
where the second equality follows from conservation of probability,
$\int \partial p/\partial\tau\,\mathrm{d}\mathbf{x}=0$.
Substituting \eqref{eq:continuity} and applying integration by parts yields
\begin{align}
    \label{eq:kl_derivative_2}
    \frac{\mathrm{d}}{\mathrm{d}\tau}\mathcal{E}_t
    & = -\int \nabla_{\mathbf{x}}\cdot(p\mathbf{v}) \log\frac{p}{q}\,\mathrm{d}\mathbf{x} \nonumber \\
    & = \int p(\mathbf{x}) \mathbf{v}(\mathbf{x})^\top \nabla_{\mathbf{x}} \log\frac{p(\mathbf{x})}{q(\mathbf{x})}\,\mathrm{d}\mathbf{x}.
\end{align}
Substituting Eqs.~(\ref{eq:relative_score}) and~(\ref{eq:sap_velocity}) into \eqref{eq:kl_derivative_2} yields a positive factor $\tilde{\omega}(t)/a_t$. Since $a_t>0$ depends only on $t$, we absorb $1/a_t$ into $\tilde{\omega}(t)$ and obtain
\begin{equation}
    \label{eq:kl_alignment}
    \frac{\mathrm{d}}{\mathrm{d}\tau}\mathcal{E}_t = -\tilde{\omega}(t) \mathbb{E}_{\mathbf{x}\sim p} \left[ \left\langle\mathbf{\Delta}, \mathcal{T}[\mathbf{\Delta}]\right\rangle \right].
\end{equation}

It remains to characterize the inner product in \eqref{eq:kl_alignment}. For each sample, \eqref{eq:purify} gives
\begin{equation}
    \label{eq:sap_fourier}
    \mathcal{F} \left[ \mathcal{T}[\mathbf{\Delta}]\right ]_c(\mathbf{u}) = \kappa \left( \mathbf{u},\mathcal{A}_{\mathbf{\Delta},c}(\mathbf{u}) \right) \mathcal{F}[\mathbf{\Delta}]_c(\mathbf{u}),
\end{equation}
where $c$ denotes the channel index. Applying Parseval's identity to the spatial inner product yields
\begin{align}
    \label{eq:parseval_sap}
    \left\langle \mathbf{\Delta}, \mathcal{T}[\mathbf{\Delta}] \right\rangle
    & = \sum_c \sum_{\mathbf{u}\in\Omega} \mathcal{F}[\mathbf{\Delta}]_c(\mathbf{u})^* \mathcal{F} \left[ \mathcal{T}[\mathbf{\Delta}] \right]_c(\mathbf{u}) \nonumber \\
    & = \sum_c \sum_{\mathbf{u}\in\Omega} \kappa \left( \mathbf{u}, \mathcal{A}_{\mathbf{\Delta},c}(\mathbf{u}) \right) \left| \mathcal{F}[\mathbf{\Delta}]_c(\mathbf{u}) \right|^2 \nonumber \\
    & = \sum_c \sum_{\mathbf{u}\in\Omega} \kappa \left( \mathbf{u}, \mathcal{A}_{\mathbf{\Delta},c}(\mathbf{u}) \right) \mathcal{A}_{\mathbf{\Delta},c}^2(\mathbf{u}).
\end{align}
Here we use the unitary DFT convention for simplicity; any alternative DFT normalization only introduces a positive constant that can be absorbed into $\tilde{\omega}(t)$.

Substituting \eqref{eq:parseval_sap} into \eqref{eq:kl_alignment} gives
\begin{equation}
    \label{eq:final_kl_descent}
    \frac{\mathrm{d}}{\mathrm{d}\tau}\mathcal{E}_t = -\tilde{\omega}(t) \mathbb{E}_{\mathbf{x}\sim p} \left[ \sum_c \sum_{\mathbf{u}\in\Omega} \kappa \left( \mathbf{u}, \mathcal{A}_{\mathbf{\Delta},c}(\mathbf{u}) \right) \mathcal{A}_{\mathbf{\Delta},c}^2(\mathbf{u}) \right] \leq 0,
\end{equation}
because $\tilde{\omega}(t)>0$ and $\kappa(\mathbf{u},\cdot)>0$.

Finally, equality in \eqref{eq:final_kl_descent} requires
$\mathcal{A}_{\mathbf{\Delta},c}(\mathbf{u})=0$
for every channel and frequency almost everywhere under $p$, which by the invertibility of the DFT implies
$\mathbf{\Delta}(\mathbf{x})=0$ almost everywhere.
From \eqref{eq:relative_score},
\begin{equation}
    \nabla_{\mathbf{x}} \log\frac{p(\mathbf{x})}{q(\mathbf{x})} = 0.
\end{equation}
Hence $\log(p/q)$ is constant on the common connected support of $p$ and $q$, so that $p=Cq$ for some constant $C$. Since both are normalized probability densities, $C=1$, yielding $p=q$ almost everywhere. Conversely, $p=q$ immediately implies $\mathbf{\Delta}=0$ and therefore $\frac{\mathrm{d}}{\mathrm{d}\tau}\mathcal{E}_t=0$.

This completes the proof of Theorem~\ref{thm:kl_descent}.

\begin{figure}[t]
    \centering
    \begin{subfigure}[b]{0.49\textwidth}
        \centering
        \includegraphics[width=\textwidth]{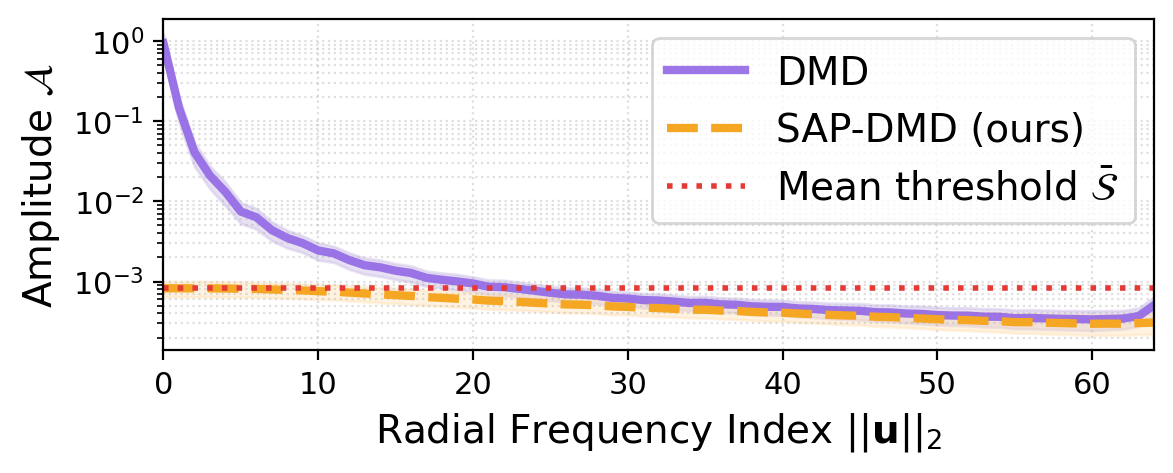}
        \caption{$t=1.0$}
    \end{subfigure}
    \hfill
    \begin{subfigure}[b]{0.49\textwidth}
        \centering
        \includegraphics[width=\textwidth]{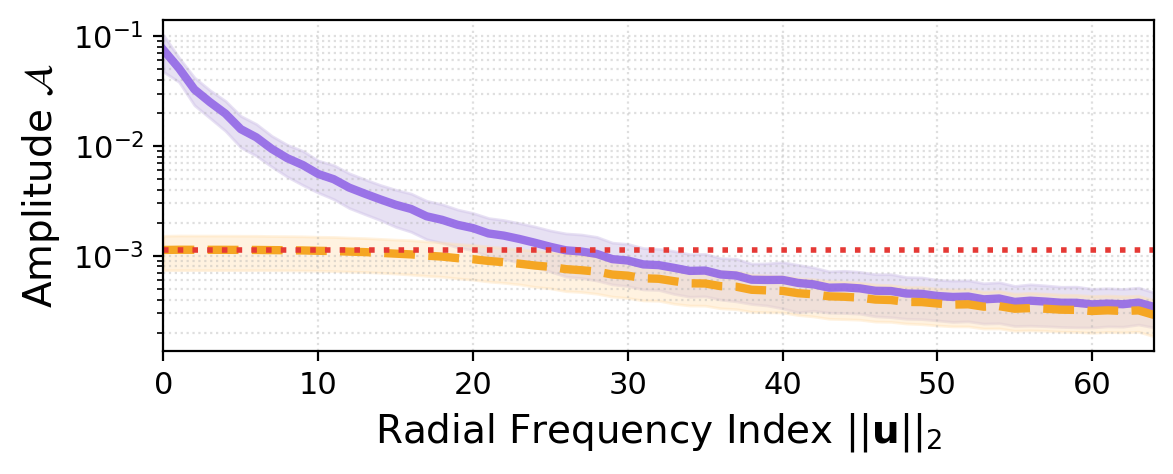}
        \caption{$t=0.75$}
    \end{subfigure}
    \hfill
    \begin{subfigure}[b]{0.49\textwidth}
        \centering
        \includegraphics[width=\textwidth]{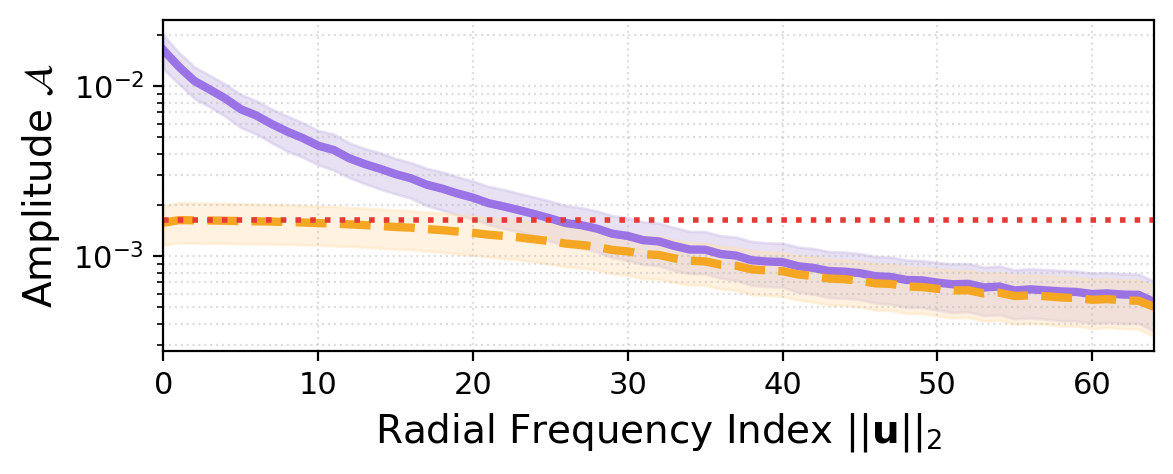}
        \caption{$t=0.50$}
    \end{subfigure}
    \hfill
    \begin{subfigure}[b]{0.49\textwidth}
        \centering
        \includegraphics[width=\textwidth]{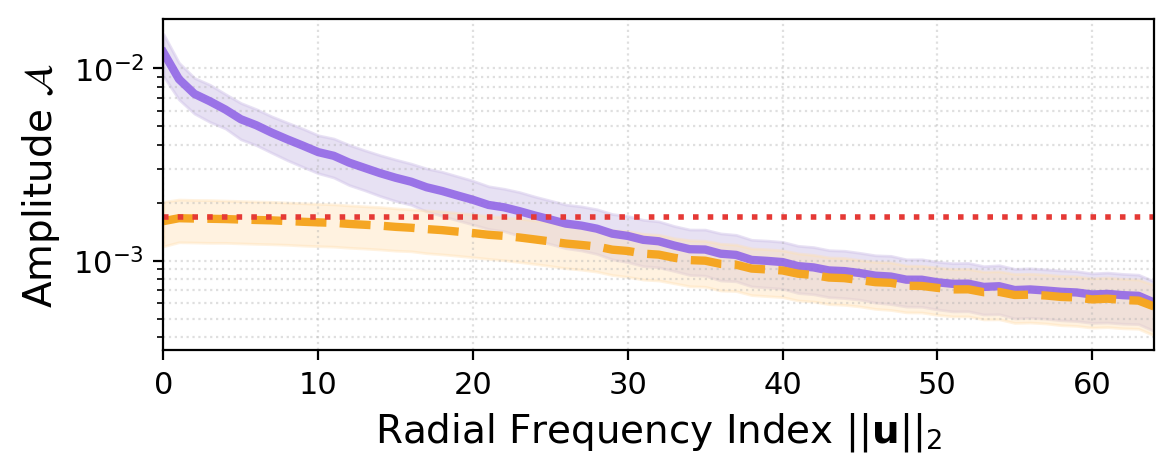}
        \caption{$t=0.25$}
    \end{subfigure}
    \caption{
    Radial amplitude profiles across representative diffusion timesteps on SD3.5 Medium. The DMD directional field consistently exhibits pronounced low-frequency amplitude concentration, while \ourName suppresses dominant amplitudes and largely preserves weaker components.
    }
    \label{fig:raps_t}
\end{figure}

\begin{figure}[t]
    \centering
    \begin{subfigure}[b]{0.49\textwidth}
        \centering
        \includegraphics[width=\textwidth]{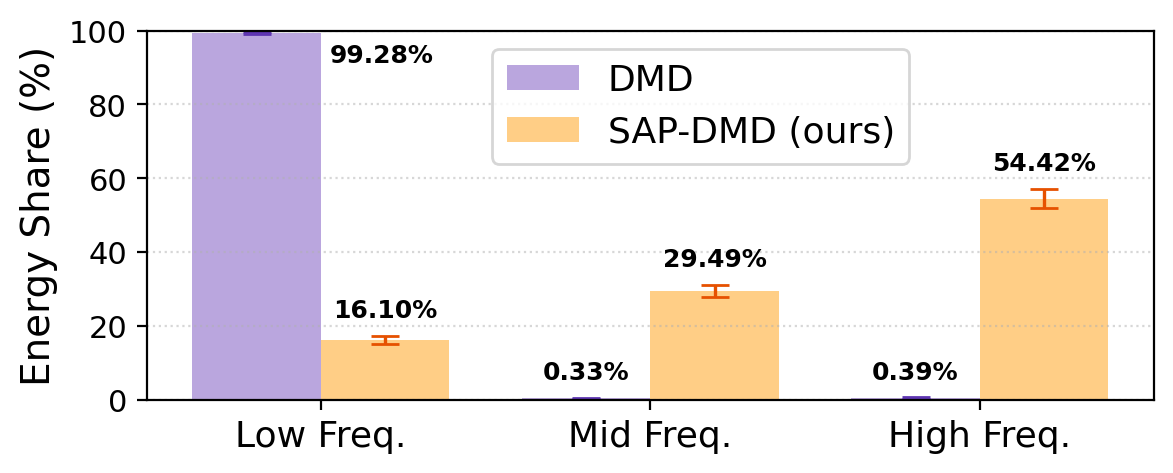}
        \caption{$t=1.0$}
    \end{subfigure}
    \hfill
    \begin{subfigure}[b]{0.49\textwidth}
        \centering
        \includegraphics[width=\textwidth]{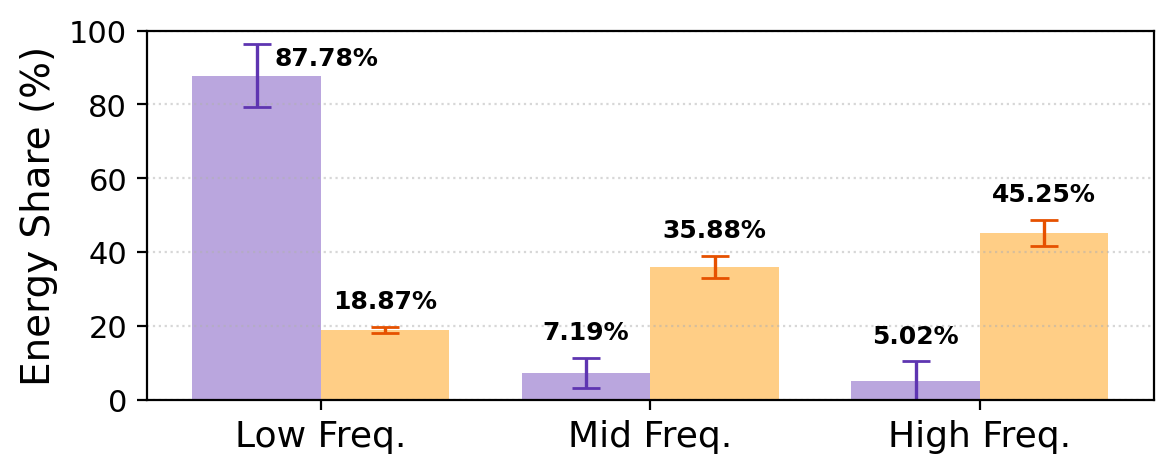}
        \caption{$t=0.75$}
    \end{subfigure}
    \hfill
    \begin{subfigure}[b]{0.49\textwidth}
        \centering
        \includegraphics[width=\textwidth]{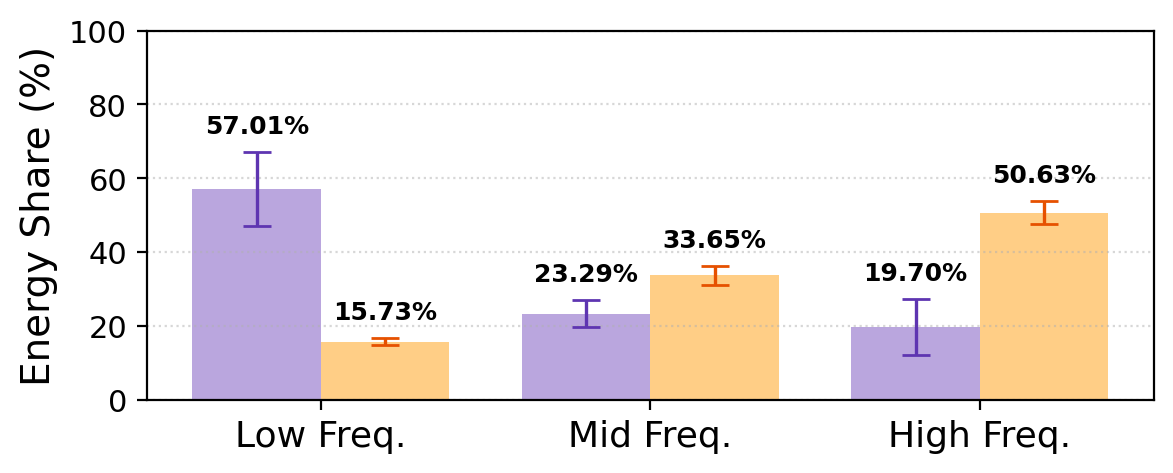}
        \caption{$t=0.50$}
    \end{subfigure}
    \hfill
    \begin{subfigure}[b]{0.49\textwidth}
        \centering
        \includegraphics[width=\textwidth]{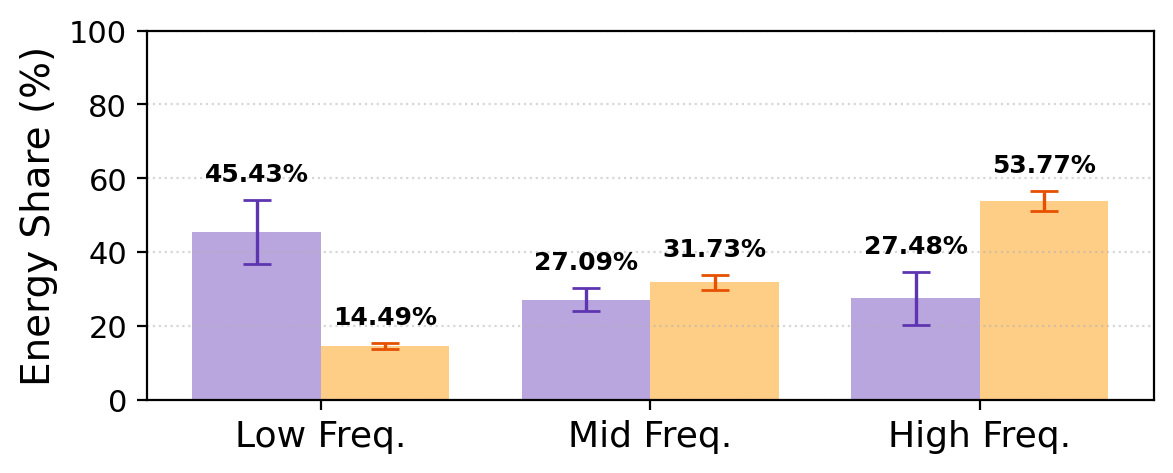}
        \caption{$t=0.25$}
    \end{subfigure}
    \caption{
    Relative spectral energy shares across representative diffusion timesteps on SD3.5 Medium. Low-frequency energy becomes increasingly dominant in the raw DMD directional field as $t$ increases. Across all timesteps, \ourName substantially reduces this relative low-frequency dominance and increases the energy shares of mid- and high-frequency components.
    }
    \label{fig:ratio_t}
\end{figure}

\begin{figure}[t]
    \centering
    \begin{subfigure}[b]{0.49\textwidth}
        \centering
        \includegraphics[width=\textwidth]{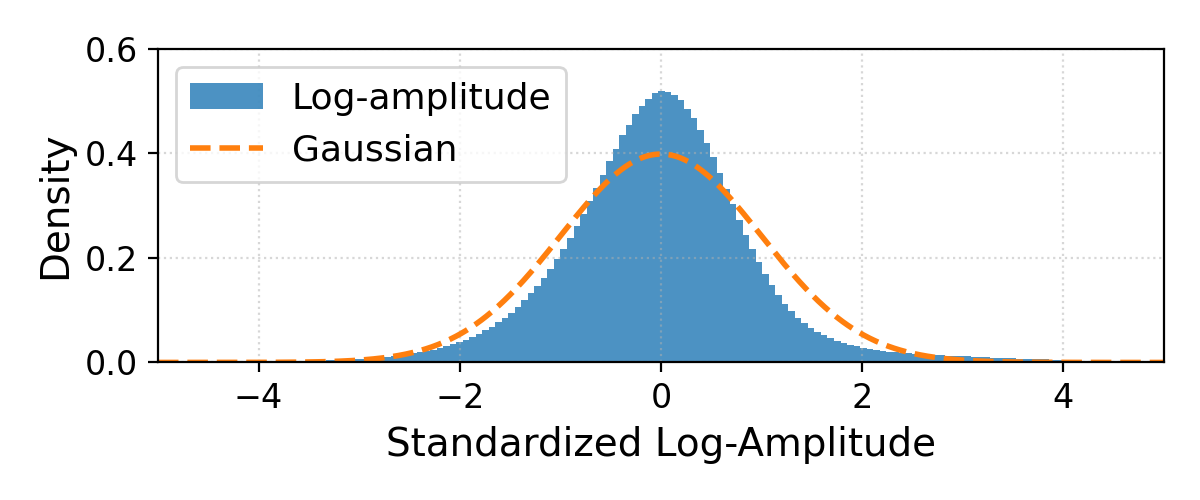}
        \caption{$t=1.0$}
    \end{subfigure}
    \hfill
    \begin{subfigure}[b]{0.49\textwidth}
        \centering
        \includegraphics[width=\textwidth]{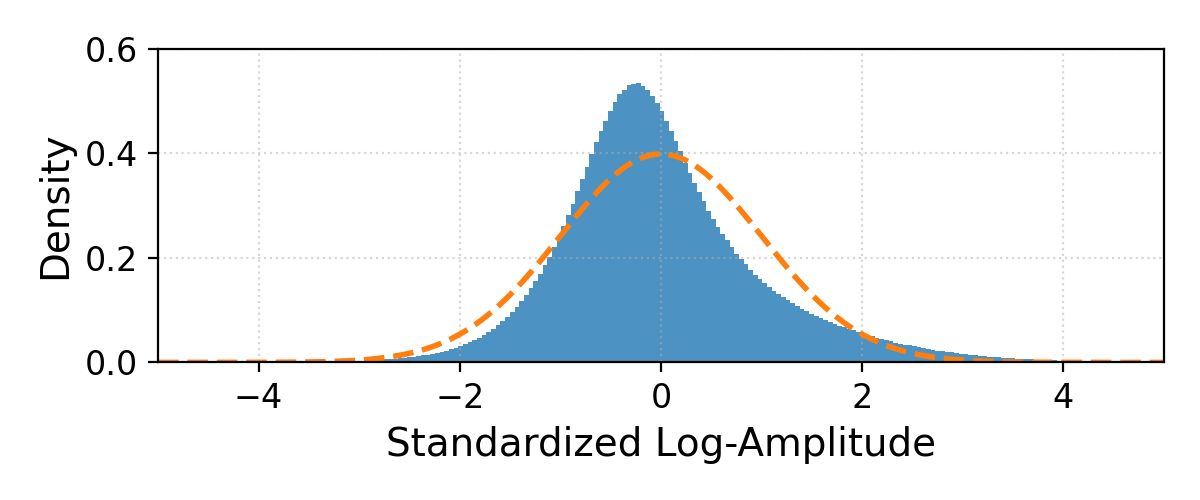}
        \caption{$t=0.75$}
    \end{subfigure}
    \hfill
    \begin{subfigure}[b]{0.49\textwidth}
        \centering
        \includegraphics[width=\textwidth]{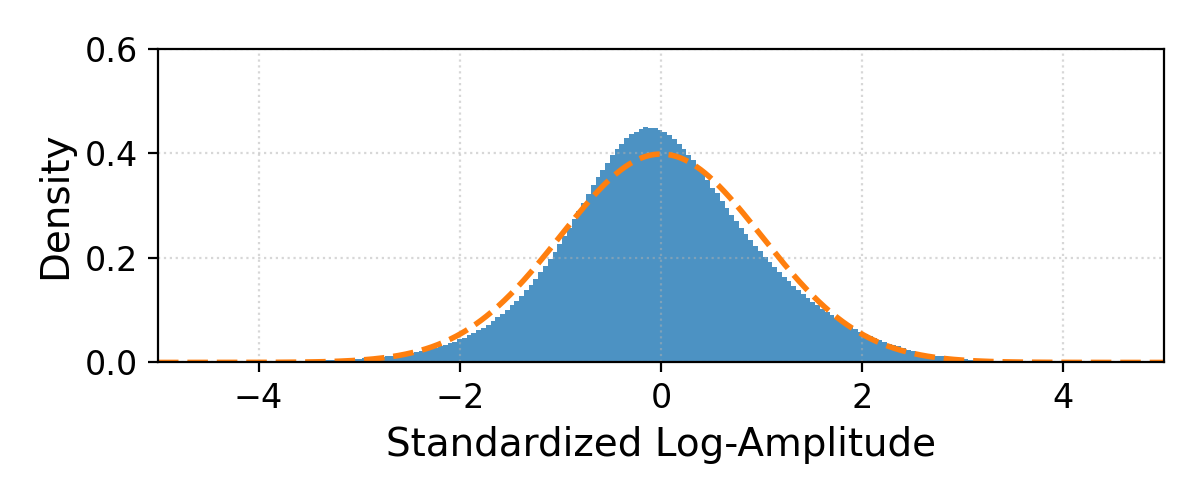}
        \caption{$t=0.50$}
    \end{subfigure}
    \hfill
    \begin{subfigure}[b]{0.49\textwidth}
        \centering
        \includegraphics[width=\textwidth]{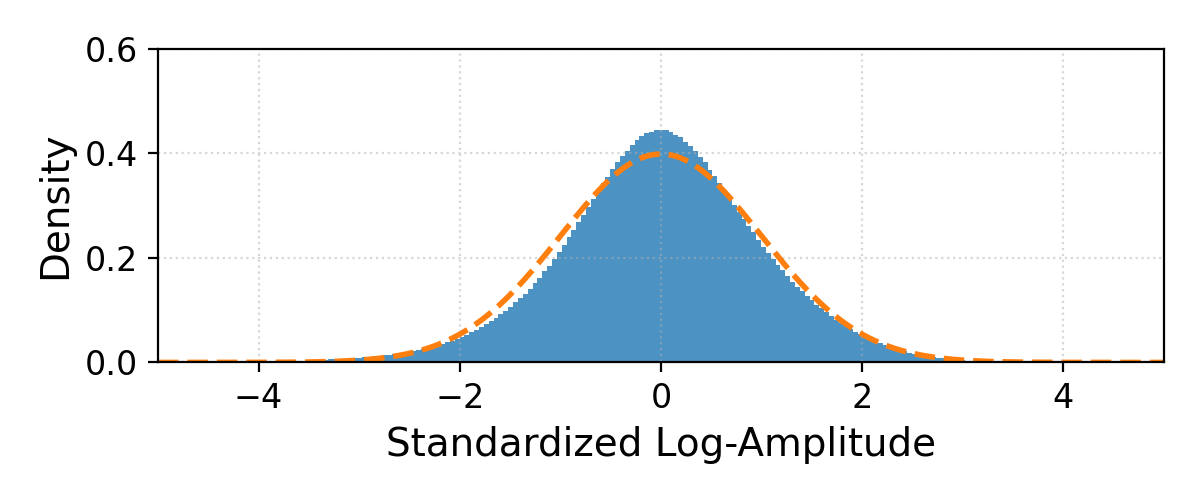}
        \caption{$t=0.25$}
    \end{subfigure}
    \caption{
    Standardized log-amplitude distributions across representative diffusion timesteps on SD3.5 Medium.
    Statistics are computed at each timestep from a DMD2 trained for 200 steps and averaged over 128 samples, with each distribution standardized independently before aggregation.
    The dashed curve denotes a standard Gaussian for reference.
    The log-amplitude distributions are substantially more symmetric than the raw amplitude distributions and remain bell-shaped across timesteps, supporting the log-domain $k$-sigma boundary used by \ourName.
    }
    \label{fig:hist_t}
\end{figure}

\section{Additional Results}

\subsection{Spectral Diagnostics Across Timesteps}
\label{app:diagnostics}
To examine whether the spectral behavior observed in Figure~\ref{fig:diag_raps} is specific to a particular diffusion timestep, we repeat the analysis at representative timesteps $t\in\{1.0,0.75,0.5,0.25\}$.
As shown in Figure~\ref{fig:raps_t}, although the absolute spectral scale varies substantially with $t$, the DMD directional field consistently exhibits low-frequency concentration.
\ourName suppresses the dominant low-frequency amplitudes across all timesteps, with weaker modulation toward higher frequencies.

The relative spectral energy statistics in Figure~\ref{fig:ratio_t} provide a complementary view.
Low-frequency dominance becomes substantially stronger at larger $t$ and gradually decreases toward smaller $t$.
\ourName consistently reduces the relative low-frequency energy share, confirming that the spectral behavior diagnosed in the main analysis is not confined to a particular diffusion timestep.

We further examine the log-amplitude statistics underlying the adaptive threshold.
As shown in Figure~\ref{fig:hist_t}, the log transform consistently produces substantially more symmetric, bell-shaped distributions across timesteps, although heavier tails remain at larger $t$.
Across 128 samples and 16 channels, the median absolute skewness decreases from $74.53$, $12.94$, $4.47$, and $3.57$ in the raw amplitude domain to $0.69$, $0.73$, $0.13$, and $0.22$ after the log transform for $t=1.0,0.75,0.5,$ and $0.25$, respectively, with similarly substantial reductions in excess kurtosis.
These results support the log domain as a more stable and symmetric space for defining the adaptive boundary $\mathcal{S}_c=\exp(\mu_c+k s_c)$.

\subsection{Propagation Through the Generator Jacobian}
\label{sec:jacobian_analysis}

SAP-DMD modifies the directional error $\mathbf{\Delta}_t$ in the output space, whereas the generator is updated by $g=J_\theta^\top\mathbf{\Delta}_t$, where $J_\theta=\partial \mathbf{x}_t/\partial\theta$. To verify that the effect of SAP-DMD survives this transformation, we decompose $\mathbf{\Delta}_t$ into low-, mid-, and high-frequency components $\mathbf{\Delta}_{t,b}$ and compute
\begin{equation}
    g_b=J_\theta^\top\mathbf{\Delta}_{t,b},
    \qquad
    q_b=
    \frac{\langle g_b,g\rangle}{\|g\|_2^2},
    \qquad
    g=\sum_b g_b.
\end{equation}
Here, $q_b$ measures the relative contribution of frequency band $b$ to the parameter update. We use the same 4,000-step DMD2 checkpoint and Jacobian for the raw and SAP-transformed signals, thereby isolating the effect of the SAP operator. Results are averaged over 128 prompts. As in the main text, the low-, mid-, and high-frequency bands are defined by $\|\mathbf{u}\|_2<16$, $16\leq\|\mathbf{u}\|_2<32$, and $32\leq\|\mathbf{u}\|_2\leq64$, respectively.

\begin{figure*}[t]
    \centering
    \includegraphics[width=0.78\textwidth]{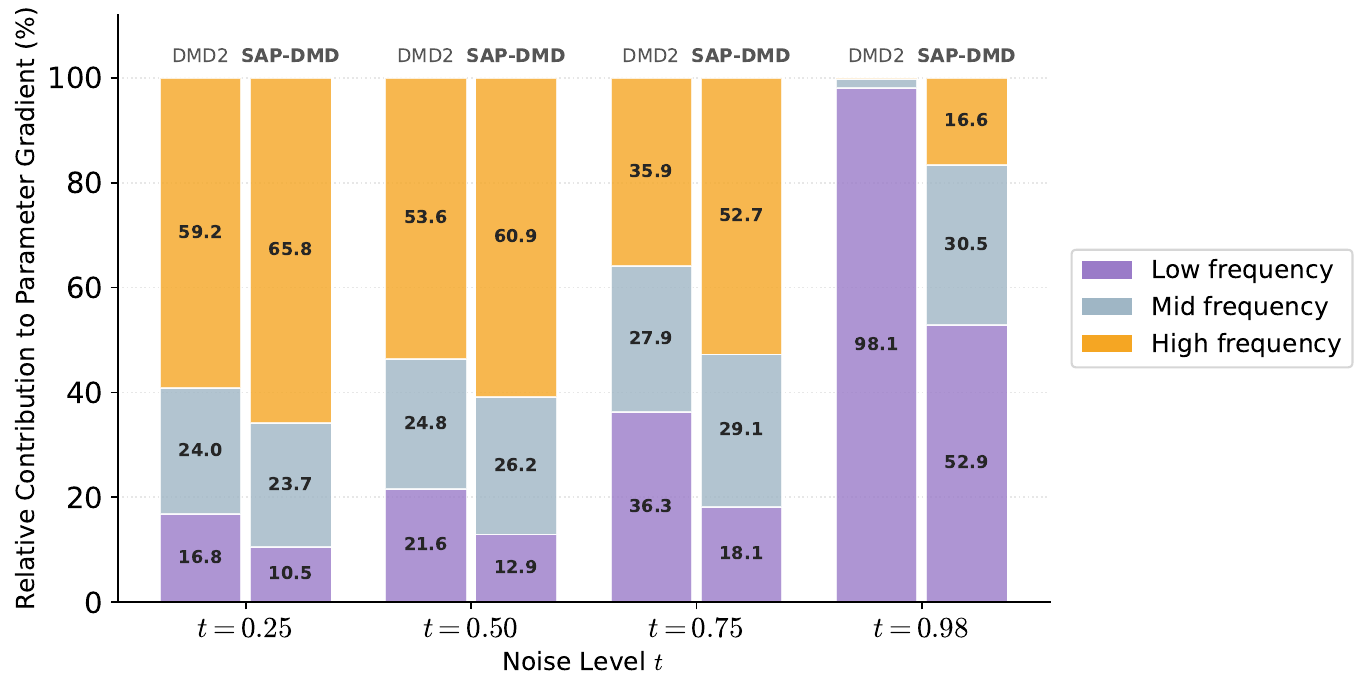}
    \caption{
    Relative contributions of different source frequency bands to the parameter gradient on SD3.5. SAP-DMD consistently reduces the low-frequency contribution and increases the high-frequency contribution after backpropagation through the generator Jacobian. $t=0.98$ is used because the generator Jacobian vanishes at the terminal timestep $t=1.0$.
    }
    \label{fig:jacobian}
\end{figure*}

As shown in Figure~\ref{fig:jacobian}, the frequency redistribution introduced by SAP-DMD remains evident after multiplication by the generator Jacobian across all tested noise levels. This confirms that SAP-DMD directly affects the parameter-space update rather than only modifying $\mathbf{\Delta}_t$ in the output space.

\subsection{Sensitivity to Training Recipe}
Our default implementation uses a resource-efficient training recipe with a global batch size of 4 and $\mathrm{TTUR}=1$, where TTUR denotes the number of fake-model updates per generator update. We examine the sensitivity of 2-step SAP-DMD on SD3.5 to these choices by varying the effective batch size and the fake-model update frequency.

\textbf{Batch size}. 
Increasing the effective batch size generally improves preference-based metrics, although the gains are modest and vary across evaluation criteria. The effect of adversarial fine-tuning also depends on the batch size: it provides clearer improvements with a larger batch, but does not consistently benefit every metric. More importantly, increasing the batch size from 4 to 32 raises the training cost by more than eightfold. Thus, while larger batches can provide moderate quality gains, they result in a substantially less favorable performance--compute trade-off. We therefore use a global batch size of 4 in our main experiments.

\textbf{TTUR}. 
Increasing the fake-to-generator update ratio from 1 to 5 provides no consistent improvement. With a batch size of 4, $\mathrm{TTUR}=5$ leaves the MS-COCO scores nearly unchanged and decreases the average HPSv2.1 score from 32.67 to 32.46, while increasing the training cost from 10.56 to 28.82 A100 hours. These results indicate that more frequent fake-model updates offer no measurable quality benefit under our setting. We therefore adopt $\mathrm{TTUR}=1$, which achieves a substantially better performance--compute trade-off.

\begin{table}[t]
    \centering
    \fontsize{8pt}{8.5pt}\selectfont
    \caption{Sensitivity of 2-step \ourName on SD3.5 Medium to the training recipe. GAN: with GAN loss in the second training stage; $+$: additional A100 hours of this stage.}
    \label{tab:app_sensitivity}
    \begin{tabularx}{\textwidth}{ccccccXXXXXc}
        \toprule
        \multirow{2}{*}{\textbf{Batch}} & \multirow{2}{*}{\textbf{TTUR}} & \multirow{2}{*}{\textbf{GAN}} & \multicolumn{3}{c}{\textbf{MS-COCO}} & \multicolumn{5}{c}{\textbf{HPSv2.1}} & \multirow{2}{*}{\textbf{A100}} \\
        \cmidrule(lr){4-6} \cmidrule(lr){7-11}
        \textbf{size} & & & \textbf{IR} & \textbf{PS} & \textbf{CS} & \textbf{Anime} & \textbf{Concept} & \textbf{Painting} & \textbf{Photo} & \textbf{Avg.} & \textbf{hours} \\
        \midrule
        4   & 1 & \multicolumn{1}{c|}{} & 1.0867 & 22.4414 & \multicolumn{1}{c|}{31.87} & 34.16 & 33.17 & 33.35 & 29.99 & \multicolumn{1}{c|}{32.67} & 10.56 \\
        4   & 5 & \multicolumn{1}{c|}{} & 1.0761 & 22.4248 & \multicolumn{1}{c|}{31.92} & 34.01 & 32.96 & 33.09 & 29.78 & \multicolumn{1}{c|}{32.46} & 28.82 \\
        4   & 1 & \multicolumn{1}{c|}{\checkmark} & 1.1104 & 22.5945 & \multicolumn{1}{c|}{32.14} & 33.58 & 32.92 & 33.04 & 29.97 & \multicolumn{1}{c|}{32.38} & +11.09 \\
        32  & 1 & \multicolumn{1}{c|}{} & 1.1245 & 22.5483 & \multicolumn{1}{c|}{32.00} & 34.28 & 33.29 & 33.32 & 30.50 & \multicolumn{1}{c|}{32.85} & 90.34 \\
        32  & 1 & \multicolumn{1}{c|}{\checkmark} & 1.1444 & 22.6080 & \multicolumn{1}{c|}{31.98} & 34.45 & 33.48 & 33.63 & 30.77 & \multicolumn{1}{c|}{33.08} & +86.39 \\
        \bottomrule
    \end{tabularx}
\end{table}

\begin{table}[t]
    \centering
    \caption{Results of 4-step DMD2 and \ourName on SD3.5 Medium without GAN loss over three matched training seeds, reported as mean $\pm$ standard deviation.
    }
    \label{tab:multiseed}
    \resizebox{\linewidth}{!}{
    \begin{tabular}{lcccccccc}
        \toprule
        \multirow{2}{*}{\textbf{Method}} & \multicolumn{3}{c}{\textbf{MS-COCO}} & \multicolumn{5}{c}{\textbf{HPSv2.1}} \\
        \cmidrule(lr){2-4}\cmidrule(lr){5-9}
        & \textbf{IR} & \textbf{PS} & \textbf{CS}
        & \textbf{Anime} & \textbf{Concept} & \textbf{Painting} & \textbf{Photo} & \textbf{Avg.} \\
        \midrule
        DMD2
        & $1.0693_{\pm .0246}$
        & $22.5196_{\pm .0359}$
        & $31.4974_{\pm .0727}$
        & $33.71_{\pm .40}$
        & $33.11_{\pm .31}$
        & $33.33_{\pm .30}$
        & $30.17_{\pm .29}$
        & $32.58_{\pm .32}$ \\
        
        SAP-DMD
        & $\mathbf{1.1125_{\pm .0009}}$
        & $\mathbf{22.6240_{\pm .0227}}$
        & $\mathbf{31.6910_{\pm .1253}}$
        & $\mathbf{34.50_{\pm .10}}$
        & $\mathbf{33.80_{\pm .10}}$
        & $\mathbf{33.98_{\pm .13}}$
        & $\mathbf{30.85_{\pm .24}}$
        & $\mathbf{33.28_{\pm .13}}$ \\
        \bottomrule
    \end{tabular}}
\end{table}

\textbf{Robustness across training seeds}.
We additionally evaluate 4-step DMD2 and SAP-DMD using three matched training seeds on SD3.5 Medium. As shown in Table~\ref{tab:multiseed}, SAP-DMD improves all metrics on average. The relatively small standard deviations further demonstrate that the improvements are reproducible across training seeds rather than being driven by a single run.

\begin{table}[t]
    \centering
    \caption{Sample diversity measured by average pairwise LPIPS (higher is more diverse) for 4-step generation. Results are reported as mean $\pm$ standard error over 128 prompts.}
    \label{tab:sample_diversity}
    \fontsize{9pt}{9pt}\selectfont
    \begin{tabular}{lcccc}
        \toprule
        Backbone & DMD2 & SAP-DMD & $\Delta$ \\
        \midrule
        PixArt-$\alpha$ & \textbf{0.568 $\pm$ 0.007} & 0.566 $\pm$ 0.006 & $-0.002$ \\
        SD3             & 0.598 $\pm$ 0.009 & \textbf{0.617 $\pm$ 0.007} & $+0.019$ \\
        SD3.5           & 0.656 $\pm$ 0.008 & \textbf{0.672 $\pm$ 0.005} & $+0.016$ \\
        \bottomrule
    \end{tabular}
\end{table}

\subsection{Sample Diversity}

We evaluate whether the quality improvements of SAP-DMD come at the cost of reduced sample diversity. Following the evaluation protocol of DMD2~\citep{yin2024improved}, we select a fixed set of 128 prompts from PartiPrompts~\citep{yu2022scaling} and generate four images per prompt using different initial noise seeds. For each prompt, we compute the LPIPS distance~\citep{zhang2018unreasonable} for all six image pairs and average the results across prompts. DMD2 and SAP-DMD use identical prompts, seeds, and four-step inference settings.

As shown in Table~\ref{tab:sample_diversity}, SAP-DMD improves pairwise LPIPS diversity on SD3 and SD3.5, while producing a statistically indistinguishable result on PixArt-$\alpha$. These results indicate that the improvements in generation quality and convergence do not arise from reduced sensitivity to the initial noise or from reduced sample diversity. Since the models generate at different native resolutions, the absolute LPIPS values should only be compared within each backbone.

\subsection{Early-Stage High-Frequency Recovery}
\label{sec:additional_detail_recovery}

\begin{wrapfigure}{r}{0.48\textwidth}
    \vspace{-1\baselineskip}
    \centering
    \includegraphics[width=\linewidth]{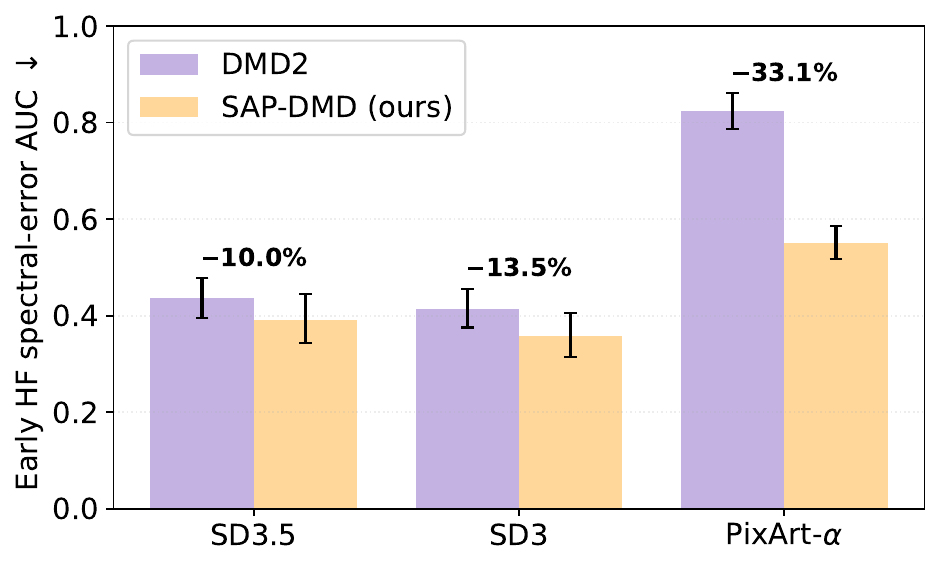}
    \caption{Early-stage high-frequency recovery on SD3.5, SD3, and
    PixArt-$\alpha$. We report the normalized AUC of the high-frequency log-power discrepancy relative to paired 25-step teacher outputs over the first $25\%$ of training. Bars show the mean over 128 prompts, and error bars denote $95\%$ bootstrap confidence intervals.
    }
    \label{fig:early_hf_auc}
\end{wrapfigure}

We quantitatively evaluate how rapidly the high-frequency spectrum of each student output approaches that of its paired 25-step teacher output during training. At each checkpoint $s$, we compute the RMS log-power discrepancy over the upper half of the normalized radial-frequency range:
\begin{equation}
    e_s =
    \sqrt{
        \frac{1}{|\mathcal{H}|}
        \sum_{r\in\mathcal{H}}
        \left(
            \log_{10} P_s(r)
            -
            \log_{10} P_{\mathrm{teacher}}(r)
        \right)^2
    },
    \qquad
    \mathcal{H}=\{r:0.5\leq r<1.0\},
\end{equation}
where $P_s(r)$ and $P_{\mathrm{teacher}}(r)$ denote the radially averaged power spectra of the student output at checkpoint $s$ and of the teacher output, respectively.
The teacher-relative discrepancy measures high-frequency recovery without rewarding arbitrary noise amplification or excessive sharpening. We summarize early-stage recovery using the normalized area under the error curve over the first $25\%$ of the 4,000-step training budget:
\begin{equation}
    \mathrm{AUC}_{\mathrm{HF}}
    =
    \frac{1}{s_K-s_1}
    \int_{s_1}^{s_K} e_s\,\mathrm{d}s,
    \qquad s_K \leq 1000,
\end{equation}
where $s_1$ and $s_K$ are the first and last evaluated checkpoints, and a lower value indicates faster convergence toward the teacher's high-frequency spectrum. As shown in Figure~\ref{fig:early_hf_auc}, SAP-DMD reduces $\mathrm{AUC}_{\mathrm{HF}}$ by $10.0\%$, $13.5\%$, and $33.1\%$ on SD3.5, SD3, and PixArt-$\alpha$, respectively. These improvements show that SAP-DMD accelerates high-frequency recovery across architectures.

\begin{figure}[t]
    \centering
    \includegraphics[width=\textwidth]{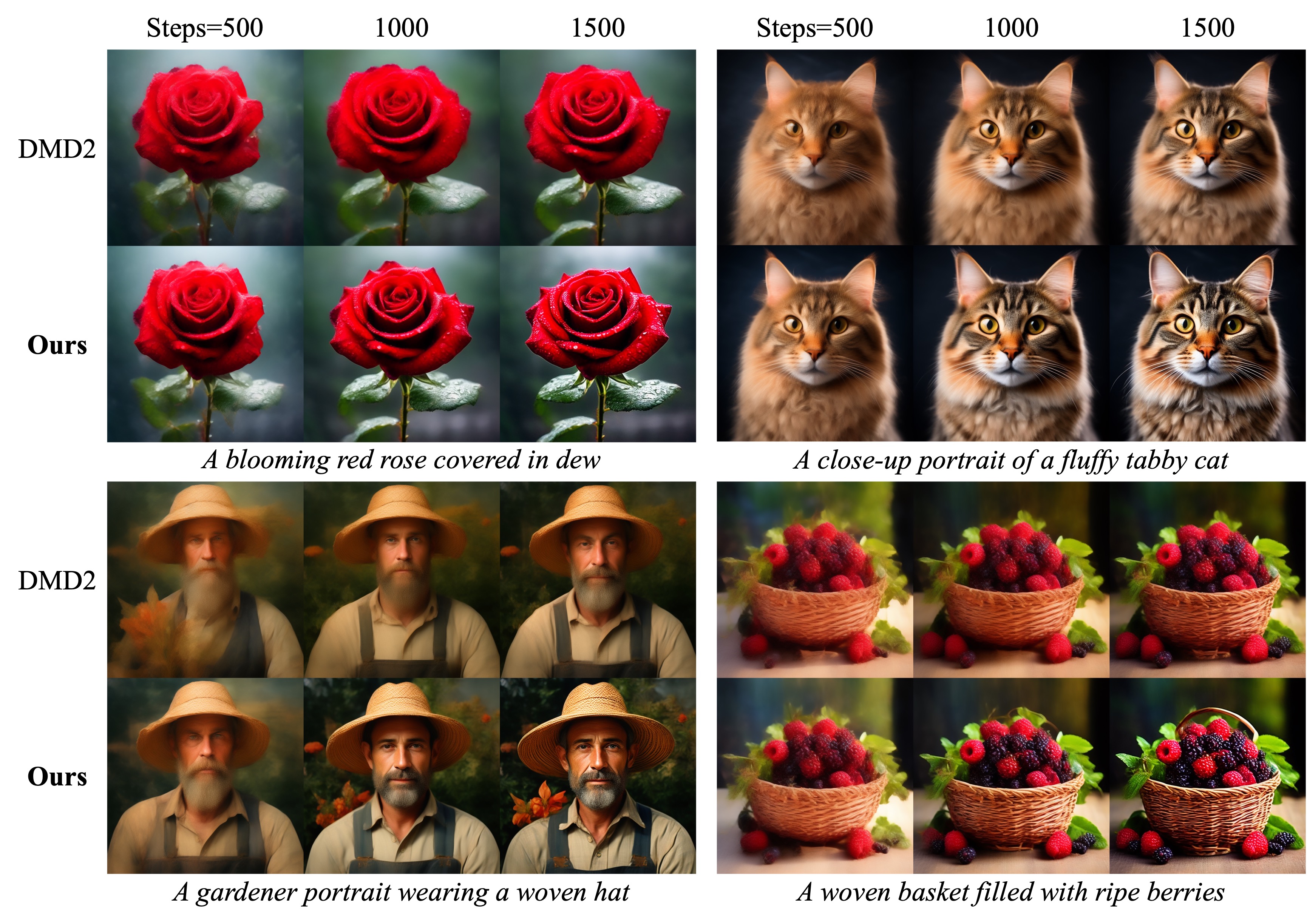}
    \caption{Fine-detail recovery during training on PixArt-$\alpha$ with 4-step generation. Columns: training steps; both methods share the prompt and initial noise.}
    \label{fig:app_evolve_pixart}
\end{figure}

\begin{figure}[t]
    \centering
    \includegraphics[width=\textwidth]{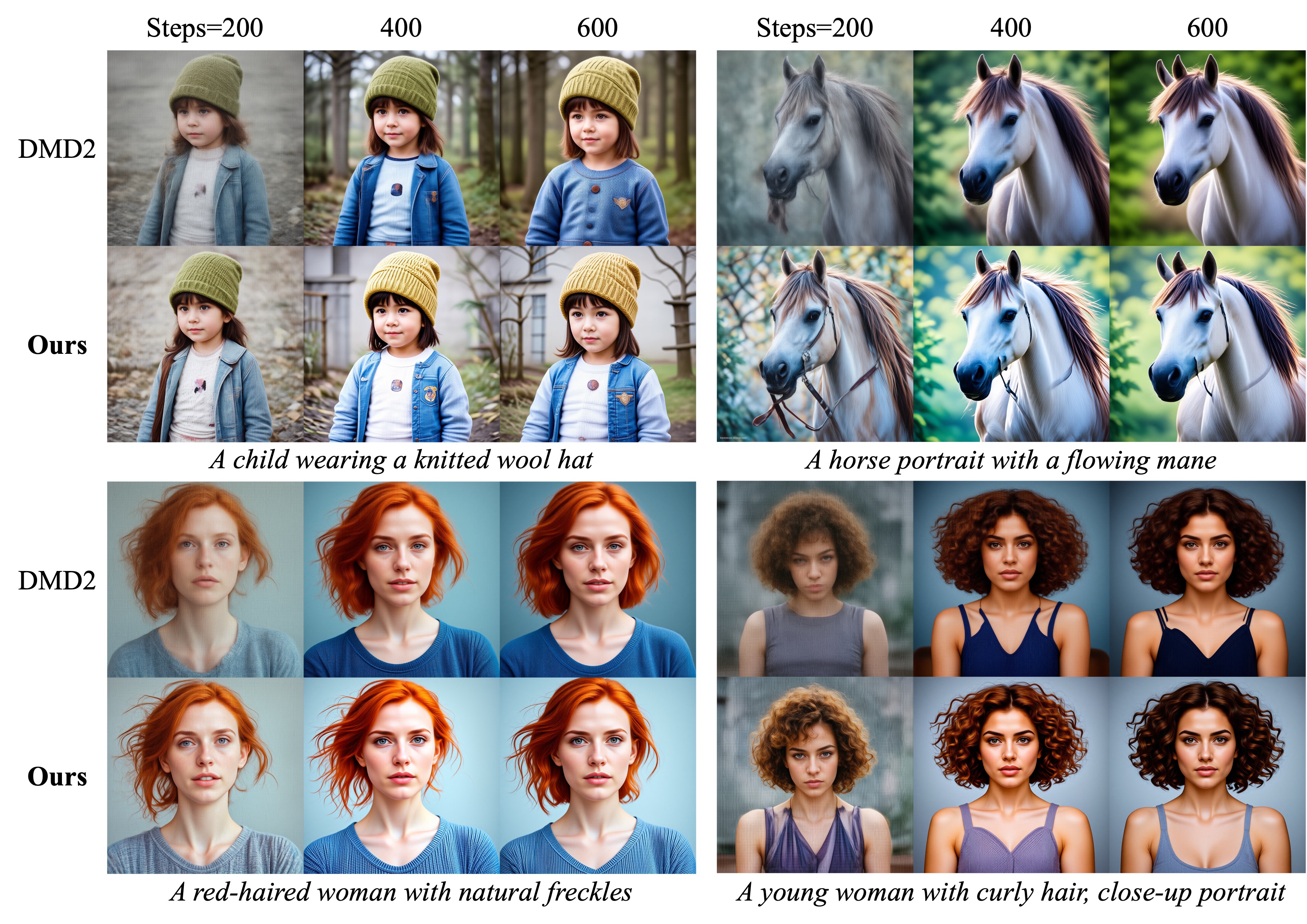}
    \caption{Fine-detail recovery during training on SD3 with 4-step generation. Columns: training steps; both methods share the prompt and initial noise.}
    \label{fig:app_evolve_sd3}
\end{figure}

\begin{figure}[t]
    \centering
    \includegraphics[width=\textwidth]{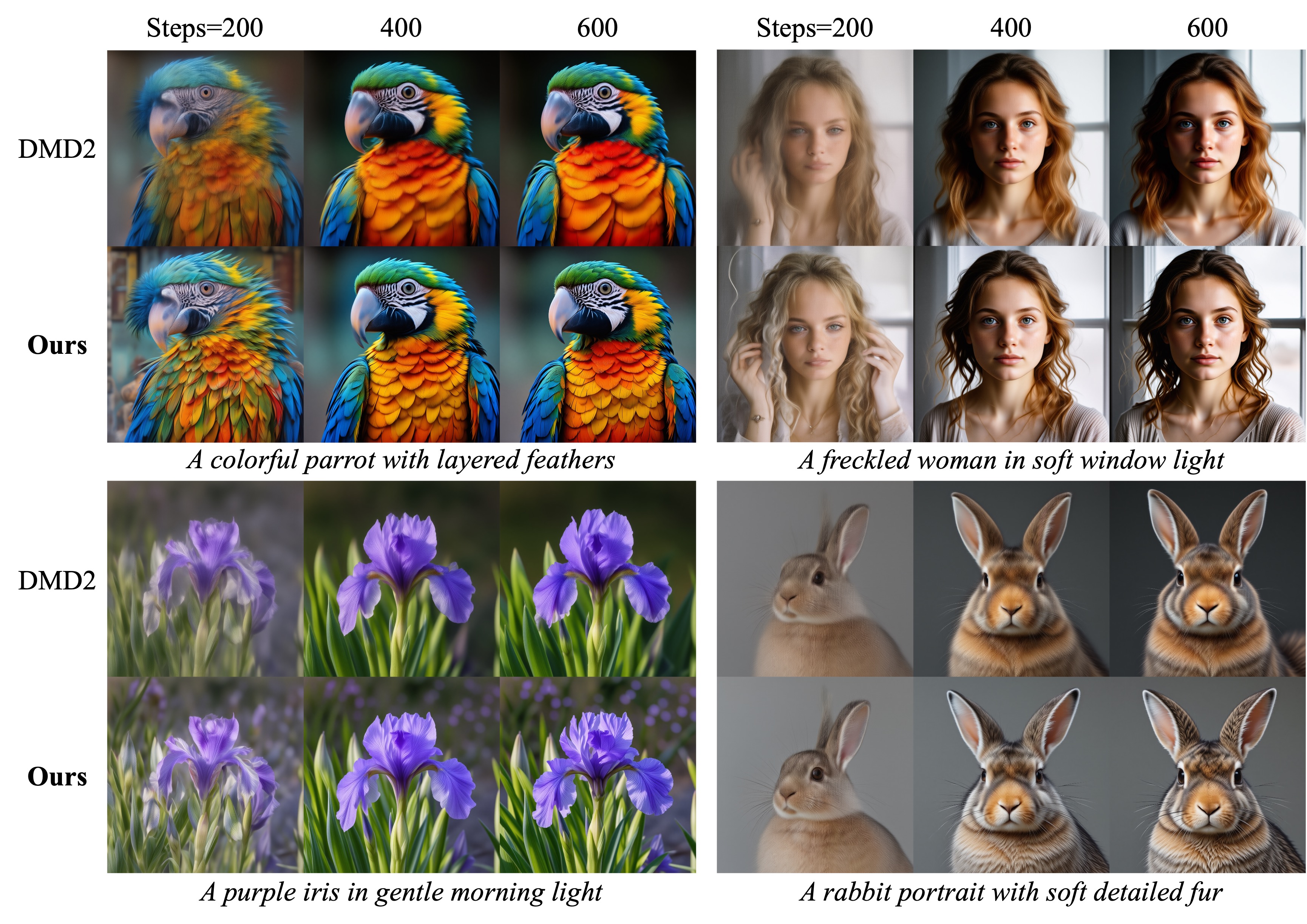}
    \caption{Fine-detail recovery during training on SD3.5 with 4-step generation. Columns: training steps; both methods share the prompt and initial noise.}
    \label{fig:app_evolve_sd35}
\end{figure}

Figures~\ref{fig:app_evolve_pixart}, \ref{fig:app_evolve_sd3}, and \ref{fig:app_evolve_sd35} provide qualitative visualizations of this recovery process using four-step generation. For each comparison, DMD2 and SAP-DMD use the same prompt, initial noise, and inference schedule. Both methods recover the coarse image structure at early checkpoints, whereas SAP-DMD develops fine-grained content, such as fur, hair, facial features, feathers, and flower petals, more rapidly.  
Across all three architectures, the improvement is already visible at intermediate checkpoints, while the global composition and semantic content remain comparable. This indicates that the observed difference primarily reflects faster recovery of local structures rather than a change in the generated content or layout. Together, the quantitative and qualitative results demonstrate that SAP-DMD accelerates fine-detail recovery during training.

\end{document}